\documentclass[10pt,journal,compsoc]{IEEEtran}
\usepackage{cite}
\usepackage{bm}
\ifCLASSINFOpdf
\usepackage[pdftex]{graphicx}
\graphicspath{{../pdf/}{../jpeg/}}
\DeclareGraphicsExtensions{.pdf,.jpeg,.png}
\else
\usepackage[dvips]{graphicx}
\graphicspath{{../eps/}}
\DeclareGraphicsExtensions{.eps}
\fi
\usepackage{amsmath}
\usepackage{amssymb}

\ifCLASSOPTIONcompsoc
\usepackage[caption=false,font=normalsize,labelfont=sf,textfont=sf]{subfig}
\else
\usepackage[caption=false,font=footnotesize]{subfig}
\fi

\usepackage{multirow}
\usepackage{color}
\usepackage{bm}
\usepackage{algorithm}
\usepackage{float}
\usepackage{booktabs}

\usepackage{tikz}
\usetikzlibrary{arrows}
\usetikzlibrary{calc}
\usetikzlibrary{arrows.meta}

\begin{document}

	\title{Meta-Teacher For Face Anti-Spoofing}
	
	\author{Yunxiao Qin, Zitong Yu, Longbin Yan, Zezheng Wang, Chenxu Zhao, Zhen Lei,~\IEEEmembership{Senior Member,~IEEE} 
		

		\IEEEcompsocitemizethanks{
			\IEEEcompsocthanksitem Y. Qin is with Neuroscience and Intelligent Media Institute (NIMI), Communication University of China, Beijing 100024, China, also with Northwestern Polytechnical University, Xian 710072, China.
			E-mail: qyxqyx@mail.nwpu.edu.cn.
			\IEEEcompsocthanksitem Z. Yu is with Center for Machine Vision and Signal Analysis, University of Oulu, Oulu 90014, Finland. E-mail: zitong.yu@oulu.fi.
			\IEEEcompsocthanksitem	L. Yan is with Northwestern Polytechnical University, Xian 710072, China.
			E-mail: yanlongbin@mail.nwpu.edu.cn.
			\IEEEcompsocthanksitem Z. Wang is with Beijing Kwai Technology Co., Ltd, Beijing 102600, China E-mail: wangzezheng@kuaishou.com.
			\IEEEcompsocthanksitem C. Zhao is with MiningLamp Technology, Beijing 100000, China. 
			E-mail: zhaochenxu@mininglamp.com.
			\IEEEcompsocthanksitem 
			Z. Lei is with the National Laboratory of Pattern Recognition (NLPR),
			Center for Biometrics and Security Research (CBSR), Institute of Automation,
			Chinese Academy of Sciences (CASIA), Beijing 100190, China, also with
			the School of Artificial Intelligence, University of Chinese Academy of
			Sciences (UCAS), Beijing 100049, China, and also with the Centre for
			Artificial Intelligence and Robotics, Hong Kong Institute of Science \&
			Innovation, Chinese Academy of Sciences, Hong Kong. E-mail:
			zlei@nlpr.ia.ac.cn.
		}
		\thanks{Manuscript received November 21, 2020; revised  May 7, 2021; accepted June 6, 2021. (Corresponding Author: Zhen Lei.)
	}}
	
	
	\markboth{IEEE Transactions on pattern analysis and machine intelligence}
	{Qin \MakeLowercase{\textit{et al.}}: Meta-Teacher For Face Anti-Spoofing}

	\IEEEtitleabstractindextext{
		\begin{abstract}
			Face anti-spoofing (FAS) secures face recognition from presentation attacks (PAs).
			Existing FAS methods usually supervise PA detectors with handcrafted binary or pixel-wise labels.
			However, handcrafted labels may are not the most adequate way to supervise PA detectors learning sufficient and intrinsic spoofing cues.
			Instead of using the handcrafted labels, we propose a novel Meta-Teacher FAS (MT-FAS) method to train a meta-teacher for supervising PA detectors more effectively.
			The meta-teacher is trained in a bi-level optimization manner to learn the ability to supervise the PA detectors learning rich spoofing cues.
			The bi-level optimization contains two key components: 1) a lower-level training in which the meta-teacher supervises the detector's learning process on the training set; and 2) a higher-level training in which the meta-teacher's teaching performance is optimized by minimizing the detector's validation loss.
			Our meta-teacher differs significantly from existing teacher-student models because the meta-teacher is explicitly trained for better teaching the detector (student), whereas existing teachers are trained for outstanding accuracy neglecting teaching ability.
			Extensive experiments on five FAS benchmarks show that with the proposed MT-FAS, the trained meta-teacher 1) provides better-suited supervision than both handcrafted labels and existing teacher-student models; and 2) significantly improves the performances of PA detectors.
		\end{abstract}

		}
	
	\maketitle
	
	%
	\IEEEdisplaynontitleabstractindextext

	%
	\IEEEpeerreviewmaketitle

	\ifCLASSOPTIONcompsoc
	\IEEEraisesectionheading{\section{Introduction}\label{sec:introduction}}
	\else
	\section{Introduction}
	\label{sec:introduction}
	\fi
	
	\IEEEPARstart{F}{ace} recognition\cite{Deng_2019_CVPR,yi2014learning,kemelmacher2016megaface,Taigman2014} has been widely utilized in identity authentication products.
	However, face recognition is vulnerable to realistic presentation attacks (PAs), including faces printed on paper (print attack), faces replayed on digital devices (replay attack), \emph{etc.}.
	Aiming to secure face recognition systems from PAs, face anti-spoofing (FAS)\cite{Pereira2012LBP,Komulainen2014Context,Shao2017Deep,zhang2020celeba,wu2020single} technology has attracted increasing attention from both academia and industry.

	In the past two decades, both traditional handcrafted feature-based~\cite{Pereira2012LBP,Komulainen2014Context,Peixoto2011Face,Boulkenafet2017Face_SURF} and deep learning-based~\cite{Lucena2017Transfer,Xu2016Learning,Shao2017Deep,zhang2020celeba,wu2020single} methods have been shown to be effective for FAS.
	On the one hand, classical handcrafted descriptors\cite{Pereira2012LBP,Komulainen2014Context,Peixoto2011Face,Boulkenafet2017Face_SURF} extract discrimination between live and spoof faces based on human prior knowledge.
	These approaches are efficient but unreliable in complex and unseen scenarios.
	On the other hand, deep learning-based methods\cite{Lucena2017Transfer,Xu2016Learning,Shao2017Deep} usually train robust presentation attack (PA) detectors to mine intrinsic spoofing patterns in an end-to-end fashion.
	Compared with handcrafted feature-based detectors, deep learning-based PA detectors usually have stronger representation capacity to detect spoof faces due to both deep networks and large-scale training data.
	
	Generally, FAS can be treated as a binary classification problem (i.e., live as `0' vs. spoof as `1'); thus, binary label with binary cross-entropy loss is widely used for supervising the PA detector. However, deep models with binary loss might discover arbitrary cues\cite{Liu2018Learning} that can separate the two classes (e.g., screen bezel) but not the faithful spoofing patterns.
	
	Recently, several hand-designed pixel-wise labels, including facial depth label\cite{Liu2018Learning,zezheng2020deep,qin2019learning}, facial reflection label\cite{kim2019basn,yu2020face}, and pixel-wise binary label\cite{DTN,george2019deep,qin2020one}, have become more popular than binary classification label in FAS.
	These pixel-wise labels utilize human's prior-knowledge about spoof faces to supervise the PA detector to learn the spoofing cues of the global distinction between
	1) live and spoof facial depths;
	2) live and spoof facial light reflections;
	and 3) live and spoof facial skin and materials (\emph{e.g.}, paper, screen), respectively.
	The facial depth label is the most popularly employed pixel-wise label.
	Because obtaining the real facial depths of all live and spoof faces is impractical, the facial depth-based FAS methods\cite{Atoum2018Face,Liu2018Learning,zezheng2020deep,yu2021revisiting} commonly train the PA detector to regress spoof faces as zero-map (each pixel value is zero) and to regress live faces as pseudo facial depths.
	When testing, they classify each face by comparing the average value of the detector's prediction map with the threshold.

	\begin{figure}[t]
		\centering
		\includegraphics[width=0.9\columnwidth]{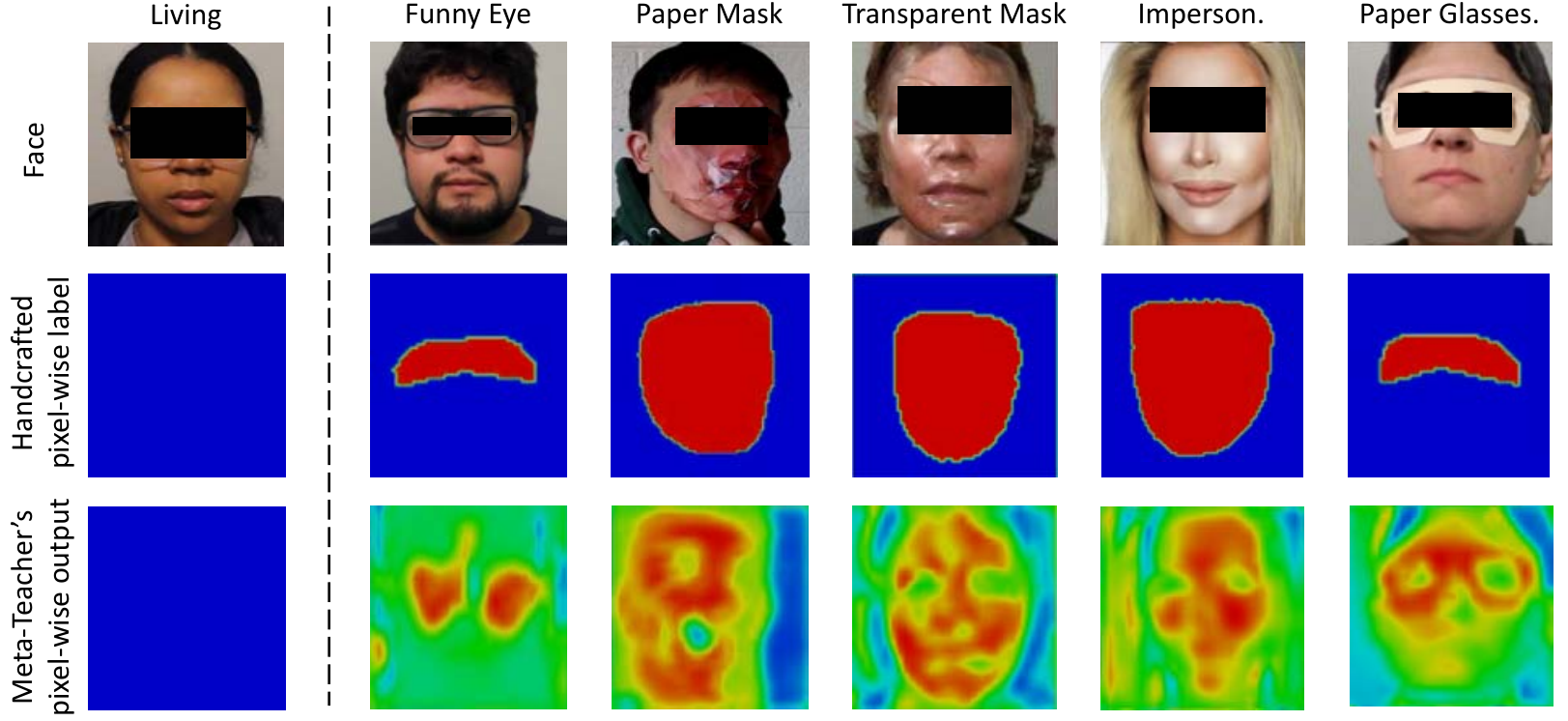}
		\vspace{-5pt}
		\caption{
			The first row: some examples of the live and spoof (funny eye, paper mask, transparent mask, impersonate, and paper glasses) faces from SiW-M\cite{DTN}.
			The second row: the exhausted annotated pixel-wise mask labels for the live and spoof faces in the first row.
			The third row: the normalized pixel-wise supervision provided by the proposed meta-teacher.
			The colors ranging from blue to red denote the float numbers ranging from 0 to 1.
			Compared with handcrafted labels, the meta-teacher can explore reasonable pixel-wise supervision to train the PA detector without using human effort.
		}
		\label{fig:fig1}
	\end{figure}
	
	Although existing handcrafted pixel-wise labels provide reasonable supervision signals for the PA detector, they might still be sub-optimal in two aspects:
	1) the design of these labels is empirical, and the human prior-knowledge about FAS applied in these labels may block the detector from exploring a broad range of spoofing cues; and 2) it is difficult for a specific human-defined label to be effective against all possible spoof types.
	As increasingly challenging attack manners are developing, the inherent global spoofing cues that these labels provide to the detector may lose effectiveness.
	For instance, as shown in Fig.~\ref{fig:fig1}, SiW-M\cite{DTN} proposed several novel spoof types, such as funny eye, paper mask, and transparent mask\footnote{Black-box-masks on the faces from SiW-M are not spoofing cues but are only used to protect personal privacy}.
	The spoofing cues are mainly located on the eyeglasses and mask; as a result, existing pixel-wise labels (facial depth label and pixel-wise binary label) may lose their effectiveness in supervising the PA detector to discriminate between live faces and these spoof faces.
	To better guide the detector in capturing the new local spoofing cues, SiW-M provides novel expensive human-annotated pixel-wise mask labels, as shown in Fig.~\ref{fig:fig1}.

	Considering the aforementioned two underlying weaknesses of human-defined labels, \textbf{we explore better pixel-wise supervision for PA detectors from a novel perspective}.
	In the field of computer vision, as an alternative to handcrafted labels, another popular and effective supervision approach is to use a well-trained teacher model to supervise the training of another deep model (student)\cite{hinton2015distilling,romero2015fitnets,furlanello2018born,jin2019knowledge,ba2014deep,mirzadeh2019improved,chen2017learning,Chen_Mei_Wang_Feng_Chen_2020}.
	This kind of method is commonly referred to as teacher-student method.
	They usually first train a powerful and larger teacher model to learn the training data.
	Then, they use the well-performing teacher model to supervise the learning of a shallower and lightweight student model.

	In this paper, inspired by teacher-student methods, we develop a novel \textbf{Meta-Teacher FAS (MT-FAS)} method to train a novel teacher called meta-teacher, to explore better pixel-wise supervision for PA detectors.
	In this work, we use the meta-teacher's prediction to supervise PA detectors.
	Our meta-teacher differs considerably from existing teachers in the following two aspects:
	
	\textbf{First}, existing teacher-student methods\cite{romero2015fitnets,jin2019knowledge,ba2014deep,furlanello2018born} do not use explicit supervision to optimize the teacher's teaching ability.
	Thus, we cannot ensure that the teacher who well matches the training data can always perform well in supervising the student\cite{cho2019efficacy} because matching the training data and supervising the student are two different tasks.
	{In contrast, the proposed MT-FAS trains the meta-teacher \textbf{exploring how to better teach (supervise) the detector (student) instead of learning the training data}.
	In other words, the optimizing objective of MT-FAS is the meta-teacher's pixel-wise supervision towards the detector (student) but not the meta-teacher's accuracy on the training set. }
	\textbf{Second}, to effectively supervise students, existing teachers are usually built with deeper and heavier models to ensure their excellent performances\cite{hinton2015distilling,romero2015fitnets}.
	In contrast, our meta-teacher does not have to be deep and heavy because it is explicitly trained to learn how to better supervise the PA detector.
	In terms of performance, we guarantee it to be perfect by manually separating its pixel-wise predictions for live and spoof faces into two different scopes.
	Specifically, we constrain each pixel value of the meta-teacher's output map for live faces to zero and constrain that for spoof faces to the range from zero to a certain positive number.
	Fig.~\ref{fig:fig1} visualizes the meta-teacher's pixel-wise predictions for live and spoof faces.
	Clearly, for spoof faces, the proposed meta-teacher has learned how to provide efficient pixel-wise supervision to train the PA detector learning the effective spoofing cues for the novel spoof types.

	\begin{figure}[t]
		\centering
		\includegraphics[width=0.7\columnwidth]{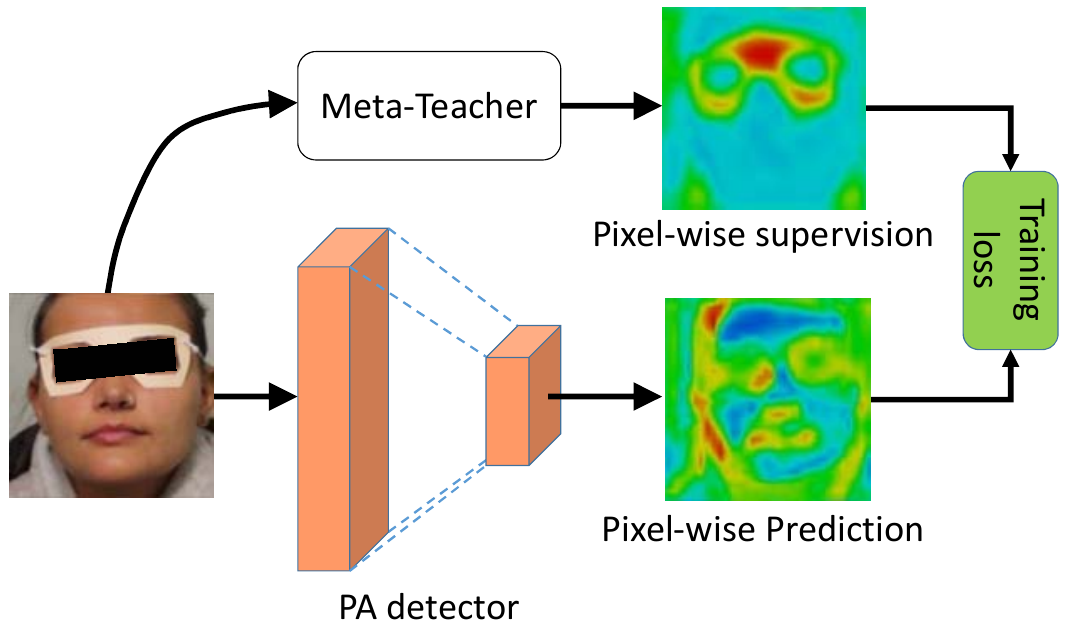}
		\vspace{-5pt}
		\caption{
			Using meta-teacher to supervise the PA detector's learning.
		}
		\label{fig:teacher_student}
	\end{figure}
	
	We utilize a bi-level optimization manner\cite{liu2018darts,MAML,qin2020layer,Meta-SGD} to train the meta-teacher.
	{In the lower-level, the meta-teacher supervises the PA detector to learn the training set.
	Normally, the more precisely the meta-teacher supervises the PA detector, the better the trained detector performs.}
	Therefore, in the higher-level, we validate the meta-teacher's teaching performance by evaluating the trained detector on the validation set.
	In this work, we use the detector's validation performance to represent the meta-teacher's teaching performance. 
	Finally, we use the detector's validation loss to optimize the meta-teacher to learn how to provide superior supervision to improve the detector's study of spoofing cues.
	After optimizing the meta-teacher, we use it to supervise existing PA detectors' learning without using human-defined labels, as illustrated in Fig.~\ref{fig:teacher_student}. 
	To sum up, the main contributions of this paper are listed:

	\begin{itemize}
		\item Instead of supervising the PA detector with handcrafted labels, we propose a novel Meta-Teacher FAS (MT-FAS) method for training a meta-teacher model to provide better-suited supervision for the PA detector.
		In contrast to existing teacher-student methods, we develop a bi-level optimization manner to explicitly optimize the meta-teacher's teaching ability, which is also novel for the field of teacher-student learning.
		\item Comprehensive experiments are conducted to verify the effectiveness of the proposed MT-FAS and meta-teacher.
		{Benefiting from the bi-level optimization objective of learning how to precisely supervise the student's learning, the proposed meta-teacher has the following advantage that compared with handcrafted labels and existing teachers, it supervises existing PA detectors more effectively and improves the performances of existing PA detectors substantially.}
		With the help of the trained meta-teacher, we update state-of-the-art performances on five FAS benchmarks.
	\end{itemize}
	
	In the remainder of the paper, Section~\ref{sec:relatedwork} provides the related works of face anti-spoofing and teacher-student. Section~\ref{sec:method}  introduces the Meta-Teacher and gives details about its implementation for the FAS task.  Section~\ref{sec:experiment} provides rigorous ablation studies and evaluates the performance of the proposed MT-FAS on five benchmark datasets. Finally, a conclusion is given in Section~\ref{sec:conclusion}.

	\section{Background}
	\label{sec:relatedwork}
	\subsection{Face Anti-Spoofing}
	In recent decades, FAS technology, which protects face recognition systems from PAs, has becomes a research hotspot. Researchers traditionally utilize well-designed feature extractors, such as local binary patterns (LBPs)~\cite{boulkenafet2015face,Pereira2012LBP}, SIFT~\cite{Patel2016Secure}, speeded-up robust features (SURF)~\cite{Boulkenafet2017Face_SURF}, histogram of oriented gradients (HOG)~\cite{Komulainen2014Context}, difference of Gaussians (DoG)~\cite{Peixoto2011Face} and remote photoplethysmography (rPPG)~\cite{li2016generalized,liu20163d,lin2019face} to extract discriminative features between live faces and spoof faces. Based on the extracted discriminative features, spoof faces can be detected with a classifier.

	Recently, deep learning-based FAS methods~\cite{Lucena2017Transfer,Nagpal2018A,Li2017An,Patel2016Cross,yu2020auto,qin2020one,Jia_2020_CVPR,zhang2020celeba,zhang2020face,yu2021revisiting,cai2020drl,liu2021face,yu2020fas} outperform traditional FAS methods on several large-scale benchmarks and have become the mainstream technology in the field of FAS.
	Specifically, deep learning-based methods usually train a deep network-based PA detector to learn discriminative features between live faces and spoof faces in a data-driven manner.
	The binary classification label (\emph{e.g.}, live as `0' and spoof as `1', or vice versa) is the most widely employed label to supervise a PA detector's training.
	Inspired by the discriminations between the facial depths of live and spoof faces, the facial depth label\cite{Liu2018Learning, shao2019regularized, zezheng2020deep,yu2020multi} was recently proposed.
	The facial depth label provides fine-grained local supervision for PA detectors, and facial depth-based methods usually supervise the detectors to regress live faces as facial depths and to regress the spoof faces as zero-maps.
	
	The reflection discrepancy between real facial skin and the surface of PAs is another inherent discrimination between live faces and spoof faces.
	The methods of \cite{yu2020face} and \cite{kim2019basn} adopt off-the-shelf generated reflection map as the supervision signal and validate the effectiveness of the reflection supervision. 
	However, this pseudo reflection label is easily influenced by environmental illumination.
	
	In addition to the above-mentioned labels, some methods\cite{DTN,george2019deep} have shown that pixel-wise binary label also works promisingly.
	Pixel-wise binary labels assume that the materials of physical carriers of spoof faces are consistent and can be used to discriminate between live faces and spoof faces.
	These methods usually set the live face label to a zero-map and set the spoof face label to a one-map (each pixel value of the map is one) or vice versa.
	
	Although existing human-defined labels are capable of supervising PA detectors to learn reasonable spoofing cues, the best-suited form of supervision remains an open question.
	In this paper, different from existing FAS methods using human-defined labels, we explore adaptive and learnable supervision signals specifically for supervising the PA detector more precisely.

	\subsection{Teacher-Student Methods}
	Traditionally, the training of neural network models is supervised by handcrafted labels.
	Several recent studies\cite{bu2006model,hinton2015distilling,ba2014deep,chen2017learning,mirzadeh2019improved,cho2019efficacy,jin2019knowledge,furlanello2018born,romero2015fitnets} show that using one or more well-trained deep and wide models to supervise another lighter model's training would benefit the lighter model's performance.
	This kind of training is commonly referred to as knowledge distillation (KD), as the lighter model distills knowledge from the cumbersome models.
	Because these methods simulate teachers' teaching process, where the cumbersome models act as teachers and the lighter models act as students, these methods are also called as teacher-student methods.
	
	Hinton et al.\cite{hinton2015distilling} propose the earliest teacher-student method.
	The authors utilize the teacher's output logits to supervise the student's training.
	In addition to the teacher's logits, FitNets\cite{romero2015fitnets} demonstrates that intermediate features of the teacher supervises the student more efficiently.
	Researchers often make the teacher's network much larger than the student's network to guarantee the teacher's superior capacity and performance.
	BANs\cite{furlanello2018born}, however, shows that the teacher network does not have to be larger than the student model.
	A teacher that has the same network as the student can still improve the student's learning.
	Furthermore, \cite{mirzadeh2019improved} and \cite{cho2019efficacy} demonstrate that improving the teacher's performance does not always enable the student to learn better.
	Teacher-student methods will lose efficacy when the representation ability gap between the teacher models and the student turns too large.
	
	Although the existing teacher-student methods are promising, they train the teacher to focus on learning how to perform better but not how to reliably teach the student.
	In this paper, we propose to train a FAS meta-teacher to focus on providing better-suited supervision signals for the student (PA detector).

	\section{Methodology}
	\label{sec:method}
	{The goal of MT-FAS is to train a meta-teacher to teach (supervise) PA detectors more precisely rather than learn the training data.
	In other words, the optimizing objective of MT-FAS is the meta-teacher's teaching ability.
	MT-FAS solves this objective via a bi-level optimizing manner that consists of lower-level learning and higher-level learning in each training iteration.}
	In the lower-level learning, the meta-teacher supervises the detector's learning process on the training set.
	In the higher-level learning, the trained detector is evaluated on the validation set, and the meta-teacher is optimized by minimizing the detector's validation loss.
	
	In this section, we detail the proposed MT-FAS with the following steps.
	First, we show how the meta-teacher supervises the detector's learning process in the lower-level learning.
	Second, we introduce how we evaluate the meta-teacher's teaching performance.
	Third, we detail the optimization of the meta-teacher in each training iteration.
	Finally, we consider more detailed problems in the implementation of MT-FAS.
	
	\begin{figure*}[t]
		\centering
		\includegraphics[width=1.8\columnwidth]{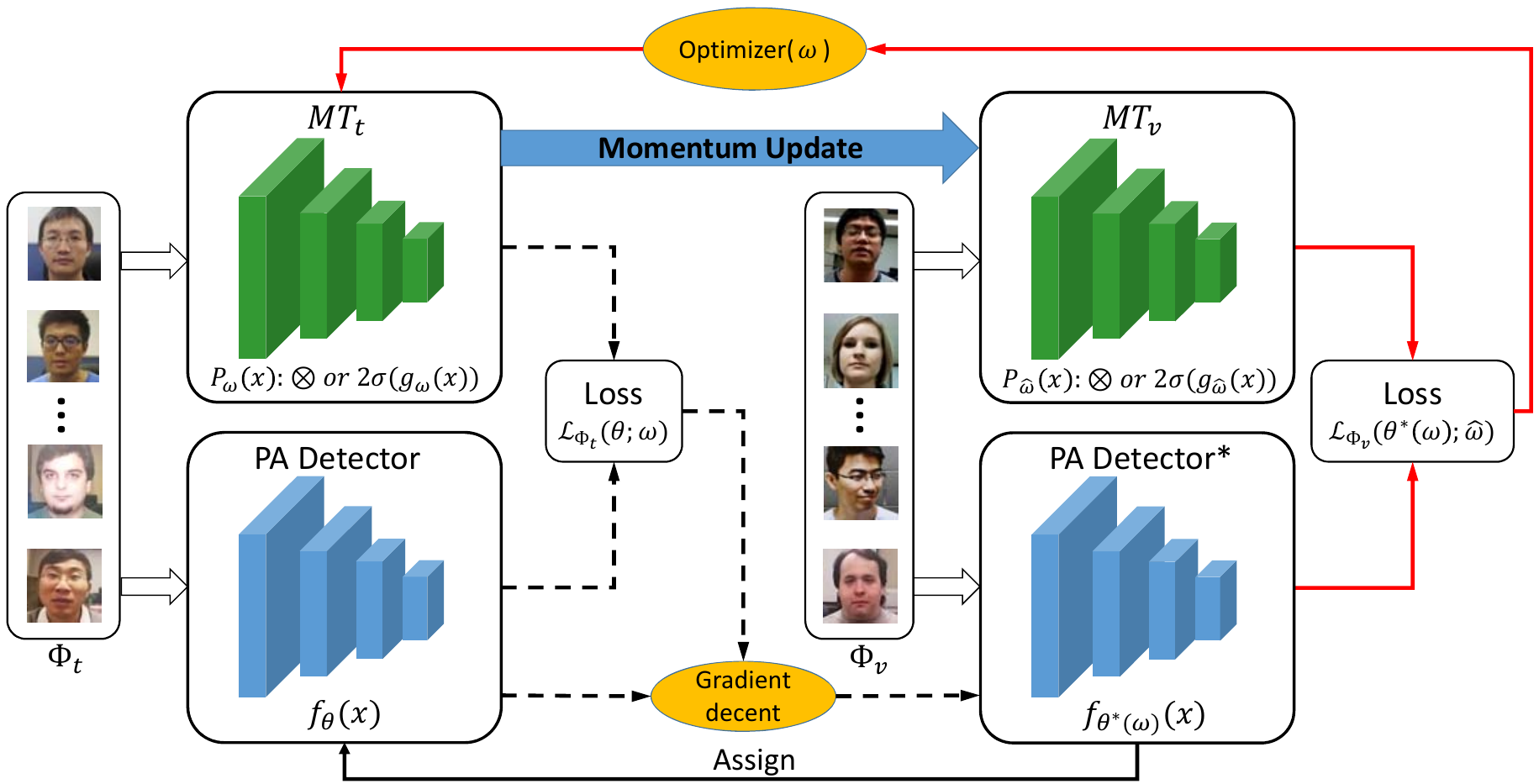}
		\caption{
			The training framework of MT-FAS.
			Each training iteration contains a lower-level and a higher-level learning.
			In the lower-level (black dotted arrow), the mini training data batch is simultaneously fed into the meta-teacher $MT_t$ and the PA detector.
			Given training data, $MT_t$ provides pixel-wise supervision for the detector.
			Be supervised by $MT_t$, the detector with weight $\theta$ is optimized to turn to newly detector with weight $\theta^*(\omega)$.
			The optimizer in the lower-level learning is gradient descent with learning rate $\alpha$.
			In the higher-level (the red arrow), the detector is evaluated on the mini validation data batch using the pixel-wise answer provided by another meta-teacher $MT_v$.
			Then, the meta-teacher $MT_t$ is optimized using the detector's validation loss $\mathcal{L}_{\Phi_v}(\theta^*(\omega)$.
			$MT_v$ is momentum updated via Eq.~\ref{eq:omega_update}.
			The PA detector's weight $\theta$ is updated by copying the weight $\theta^*(\omega)$, as the black arrow shows.
		}
		\label{fig:framework}
	\end{figure*}
	
	\subsection{Using the Meta-Teacher to Supervise the Detector}
	\label{Sec:inner-up}
	In the lower-level learning of each bi-level training iteration, we use the meta-teacher to supervise the PA detector (student) to learn the training set.
	The learning of the detector can be formulated as 
	\begin{equation}
	\begin{aligned}
	\mathcal{L}_{\Phi_t}(\theta; \omega) &= \frac{1}{N_t} \sum_{k}^{N_t} \|f_{\theta}(x_k) - P_{\omega}(x_k)\|_2,   \\
	\theta^*(\omega) &= \mathop{argmin}\limits_{\theta} \mathcal{L}_{\Phi_t}(\theta; \omega) ,
	\end{aligned}
	\label{eq:loss_MSE_train}
	\end{equation}
	where $\Phi_t$ is the training set and $x_k$ is the $k$-th training face;
	$N_t$ is the number of faces in the training set;
	$f_{\theta}$ is the detector parameterized by $\theta$;
	$P_{\omega}$ is the meta-teacher parameterized by $\omega$.
	In this work, either the meta-teacher or the detector predicts each input face as a map since numerous existing works\cite{Liu2018Learning,zezheng2020deep,qin2019learning,kim2019basn,yu2020face,qin2020one} demonstrate that pixel-wise regression supervision outperforms binary cross-entropy supervision.
	Given face $x_k$, $P_{\omega}(x_k)$ and $f_{\theta}(x_k)$ are the meta-teacher and the detector's output maps, respectively.
	$\mathcal{L}_{\Phi_t}(\theta)$ is the detector's mean-square error (MSE) loss on the training set. 
	
	In the lower-level learning, MT-FAS uses the meta-teacher's output map to supervise the detector's training on the training set.
	As $\theta$ is optimized with the supervision provided by the meta-teacher, we use $\theta^*(\omega)$ to denote the weight of the optimized detector.
	$\theta$ and $\omega$ can also be understood as the lower-level and upper-level weights\cite{anandalingam1992hierarchical,colson2007an}, respectively.

	To guarantee perfect meta-teacher performance in distinguishing live and spoof faces, we manually set the output of the meta-teacher $P^t(x_k)$ to
	\begin{equation}
	P_{\omega}(x_k)=
	\left\{
	\begin{array}{lr}
	\otimes, \qquad \qquad \qquad \  if \ x_k \in live,   \\
	2\sigma(g_{\omega}(x_k)), \; \quad \quad if \ x_k \in spoof,
	\end{array}
	\right.
	\label{eq:Pt}
	\end{equation}
	where $\otimes \in \mathcal{R}^ {32 \times 32}$ is the zero-map for live faces and $2\sigma(g_{\omega}(x_k)) \in \mathcal{R}^ {32 \times 32}$ is the output pixel-wise map for spoof faces.
	$g_{\omega}$ is the network of the meta-teacher, and $\sigma$ is the sigmoid function such that each pixel value in the map $2\sigma(g_{\omega}(x_k))$ is constrained to the range of 0 to 2.

	\subsection{Evaluating the Meta-Teacher's Teaching Quality}
	\label{Sec:outer-up1}
	The higher-level learning of each bi-level training iteration evaluates and optimizes the meta-teacher's teaching performance.
	Here, we introduce how we evaluate the meta-teacher's teaching performance.
	The optimization details of the meta-teacher will be introduced in Section~\ref{Sec:outer-up2}.
	
	Theoretically, the meta-teacher's teaching quality determines the performance of the trained student (detector); that is, better teaching quality of the meta-teacher will lead to better performance of the detector on the validation set.
	Therefore, the most straightforward way to evaluate the quality of the teacher's teaching is to evaluate the performance of the trained student on the validation/testing set.
	In this work, we use the student's performance on the validation set to represent the teacher's teaching ability.
	The detector's validation loss can be formulated as
	\begin{equation}
	\begin{aligned}
	\mathcal{L}_{\Phi_v}(\theta^*(\omega); \widehat{\omega}) = \frac{1}{N_v} \sum_{k}^{N_v} \|f_{\theta^*(\omega)}(x_k) - P_{\widehat{\omega}}(x_k)\|_2,
	\end{aligned}
	\label{eq:loss_MSE_val}	
	\end{equation}
	where $\Phi_v$ denotes the validation set and $N_v$ is the number of examples in $\Phi_v$.
	$x_k$ is the $k$-th example in $\Phi_v$.
	$P_{\widehat{\omega}}$ is another meta-teacher on the validation set, and $\widehat{\omega}$ is the corresponding weight. 
	For clarity, we denote the meta-teacher $P_{\omega}$ on the training set as $MT_t$ and denote the meta-teacher $P_{\widehat{\omega}}$ on the validation set as $MT_v$.
	The reason we use two different meta-teachers is that we use the detector's validation loss to represent $MT_t$'s teaching ability.
	If the detector's validation loss is calculated with $MT_t$, then the situation in which $MT_t$ is evaluated by itself will occur.
	This situation may result in misestimation of the meta-teacher and further damage the training of the meta-teacher.
	To avoid this issue, a new $MT_v$ is needed to calculate the validation loss $\mathcal{L}_{\Phi_v}(\theta^*(\omega); \widehat{\omega})$ and evaluate $MT_t$.
	{We will present experiments in Section~\ref{sec:exp_MT_v} to demonstrate the indispensability of $MT_v$.}
	Similar to $MT_t$, we formulate $MT_v$ as
	\begin{equation}
	P_{\widehat{\omega}}(x_k)=
	\left\{
	\begin{array}{lr}
	\otimes, \qquad \qquad \qquad \  if \ x_k \in live,   \\
	2\sigma(g_{\widehat{\omega}}(x_k)), \; \quad \quad if \ x_k \in spoof,
	\end{array}
	\right.
	\label{eq:Pv}
	\end{equation}
	where $g_{\widehat{\omega}}$ is the network of $MT_v$. 
	The main difference between $MT_t$ and $MT_v$ is that they predict pixel-wise maps for spoof faces using different weights: the weight of $MT_v$ is $\widehat{\omega}$, while that of $MT_t$ is $\omega$.
	
	To ensure that the meta-teacher $MT_t$'s teaching ability is correctly evaluated, we propose the \textbf{premise} that \emph{a better $MT_t$ can teach the detector to achieve a lower validation loss}.
	To satisfy this premise, $MT_t$ and $MT_v$ should closely correlate with each other despite their differences.
	The relationship between $MT_t$ and $MT_v$ will be discussed in greater detail in Section \ref{Sec:concerns}.
	
	Typically, if the premise is perfectly satisfied, the best $MT_t$ should lead the detector to achieve the minimal validation loss.
	Therefore, the best weight $\omega^*$ for $MT_t$ can be formulated as
	\begin{equation}
	\omega^* = \mathop{argmin}\limits_{{\omega}} \mathcal{L}_{\Phi_v}(\theta^*(\omega); \widehat{\omega}).
	\label{eq:label_optimize}
	\end{equation}

	\subsection{Optimization of the Meta-Teacher}
	\label{Sec:outer-up2}
	The purpose of MT-FAS is to train the meta-teacher $MT_t$ to provide better pixel-wise supervision to train the detector $f_{\theta}$, as Eq.~\ref{eq:label_optimize} shows.
	Here, we detail how $MT_t$ is optimized.
	
	First, we approximate the optimization of $\theta^*$ with one gradient descent step, formulated as
	\begin{equation}
	\theta^*(\omega) \approx \theta - \alpha \cdot \nabla_{\theta}\mathcal{L}_{\Phi_t}(\theta; \omega),
	\label{eq:label_omega}
	\end{equation}
	where $\alpha$ is the detector's learning rate on the training set.
	Since we wish the detector trained with $P_{\omega}$ to achieve the minimal validation loss, we optimize $P_{\omega}$ by minimizing the detector's validation loss.
	The formulation of the optimization is
	\begin{equation}
	\omega' = \omega - \beta \cdot \nabla_{\omega} \mathcal{L}_{\Phi_v}(\theta^*(\omega); \widehat{\omega}),
	\label{eq:label_optimize3}
	\end{equation}
	where $\beta$ denotes the learning rate of the meta-teacher.
	By using the chain rule, we reformulate $\nabla_{\omega} \mathcal{L}_{\Phi_v}(\theta^*(\omega); \widehat{\omega})$ as
	\begin{equation}
	\nabla_{\omega} \mathcal{L}_{\Phi_v}(\theta^*(\omega); \widehat{\omega}) = \nabla_{\omega} \theta^*(\omega) \cdot \nabla_{\theta^*(\omega)} \mathcal{L}_{\Phi_v}(\theta^*(\omega); \widehat{\omega}). \\
	\label{eq:label_optimize32}
	\end{equation}
	
	We replace $\theta^*(\omega)$ using Eq.~\ref{eq:label_omega} and write the first item $\nabla_{\omega} \theta^*(\omega)$ in detail as
	\begin{equation}
	\begin{aligned}
	\nabla_{\omega} \theta^*(\omega) &\approx  \nabla_{\omega} (\theta - \alpha \cdot \nabla_{\theta}\mathcal{L}_{\Phi_t}(\theta; \omega)) \\
	&= -\alpha \cdot \nabla_{\omega} \nabla_{\theta}\mathcal{L}_{\Phi_t}(\theta; \omega) \\
	&= -\alpha \cdot \nabla_{\omega} \nabla_{\theta}\frac{1}{N_t} \sum_{k}^{N_t} \|g_{\theta}(x_k) - P_{\omega}(x_k)\|_2.
	\end{aligned}
	\label{eq:label_optimize33}
	\end{equation}
	$\frac{1}{N_t} \sum_{k}^{N_t} \|g_{\theta}(x_k) - P_{\omega}(x_k)\|_2$ is the training loss and is differentiable with respect to $P_{\omega}(x_k)$.
	$P_{\omega}(x_k)$ is also differentiable with respect to $\omega$.
	Therefore, we can calculate the gradient $\nabla_{\omega} \nabla_{\theta}\frac{1}{N_t} \sum_{k}^{N_t} \|g_{\theta}(x_k) - P_{\omega}(x_k)\|_2$.
	Finally, we rewrite the optimization of $MT_t$ as
	\begin{equation}
	\begin{aligned}
	\omega^{'} \! \! &= \! \omega \! - \! \beta \! \cdot \! \nabla_{\omega} \mathcal{L}_{\Phi_v}(\theta^*(\omega); \widehat{\omega}) \\
	&= \! \omega \! + \! \beta \! \cdot \! \alpha \! \cdot \! \nabla_{\omega} \nabla_{\theta} \mathcal{L}_{\Phi_t}(\theta; \omega)  \! \cdot \! \nabla_{\theta^*(\omega)} \mathcal{L}_{\Phi_v}(\theta^*(\omega); \widehat{\omega}).
	\end{aligned}
	\label{eq:label_optimize5}
	\end{equation}

	\begin{algorithm}[t]
		\caption{Training of Meta-teacher}
		{\bfseries input:} FAS training set \emph{$\Psi_t$}, learning rates $\beta$ and $\alpha$, update interval $T$, momentum update parameter $\gamma$, two batch size values $M$ and $N$. \\
		{\bfseries output:} The meta-teacher's weight $\omega$. \\
		{\bfseries 1\,\,\,:} Initialize $\omega$ and $\theta$ by pretraining $MT_t$ and the detector on the training set, initialize $\widehat{\omega}$ to $\omega$. \\
		{\bfseries 2\,\,\,:} Iter = 0 \\
		{\bfseries 3\,\,\,:} {\bfseries while} not done {\bfseries do} \\
		{\bfseries 4\,\,\,:}  \ \ \,	  sample $M+N$ live and $M+N$ spoof faces from $\Psi_t$.  \\
		{\bfseries 5\,\,\,:}  \ \,build mini training data batch $\Phi_t$ with $2M$ live and spoof faces; build mini validation data batch $\Phi_v$ with the other $2N$ faces.  \\
		{\bfseries 6\,\,\,:}  \ \ \,	$\mathcal{L}_{\Phi_t}(\theta; \omega) = \frac{1}{2M} \sum_{k}^{2M} \|f_{\theta}(x_k) - P_{\omega}(x_k)\|_2$   \\
		{\bfseries 7\,\,\,:}  \ \ \,  $\theta^*(\omega) = \theta - \alpha \cdot \nabla_{\theta}\mathcal{L}_{\Phi_t}(\theta; \omega)$\\
		{\bfseries 8\,\,\,:}     \ \ \, $\mathcal{L}_{\Phi_v}(\theta^*(\omega); \widehat{\omega}) = \frac{1}{2N} \sum_{k}^{2N} \|f_{\theta^*(\omega)}(x_k) - P_{\widehat{\omega}}(x_k)\|_2$\\
		{\bfseries 9\,\,\,:} 	\ \ \, $\omega = \omega +  \beta  \cdot  [\ \alpha  \cdot  \nabla_{\omega} \nabla_{\theta} \mathcal{L}_{\Phi_t}(\theta; \omega)   \cdot  \nabla_{\theta^*(\omega)} \mathcal{L}_{\Phi_v}(\theta^*(\omega); \widehat{\omega}) + \nabla_{\omega} \frac{\mu}{2M} \sum_{k}^{2M} mean(\sigma(g_{\omega}(x_k))) \ ]$ \\
		{\bfseries 10:} 	\ \ \, $\widehat{\omega} = \gamma \cdot \widehat{\omega} + (1-\gamma) \cdot \omega$ \\
		{\bfseries 11:} \ \ \,  {\bfseries if} Iter {\bfseries \%} $T == 0$  {\bfseries do} \\				
		{\bfseries 12:} 	\ \ \ \ \ \, $\theta = \theta^*(\omega)$\\				
		{\bfseries 13:} 	\ \ \, Iter $\leftarrow$ Iter + 1\\
		{\bfseries 14:} 	{\bfseries end}
		\label{algorithm:Meta-teacher}
	\end{algorithm}

	\subsection{Other Details of the Meta-Teacher}
	\label{Sec:concerns}
	From Eq.~\ref{eq:label_optimize5}, we can see how $MT_t$ is optimized.
	We present the full training procedure of MT-FAS in Algorithm~\ref{algorithm:Meta-teacher}.
	Here, we discuss some implementation details.

	\subsubsection{Initialization of $MT_t$, $MT_v$, and the detector} 
	Before optimizing $MT_t$, we first initialize $MT_t$, $MT_v$, and the detector to appropriate weights by pretraining them on the training set according to the following reasons: 
	i) randomly initialized $MT_t$ and $MT_v$ output noisy maps for spoof faces;
	ii) a randomly initialized detector predicts noisy maps for all input faces.
	These factors may lead to a failure to satisfy the aforementioned premise.
	In the pretraining step, we train $MT_t$ and the detector to regress all spoof faces as one-maps (each pixel value is one) and to regress live faces as zero-maps.
	After pretraining of $MT_t$, we initialize $MT_v$ with the pretrained $MT_t$.
	Line 1 of Algorithm~\ref{algorithm:Meta-teacher} shows the pretraining.
	Experiments in Section~\ref{sec:pretrain} verify the importance of pretraining.

	\subsubsection{Relationship between $MT_t$ and $MT_v$}
	According to Eq.~\ref{eq:loss_MSE_val}, we use another meta-teacher $MT_v$ to calculate the trained detector's validation loss and use the validation loss to represent the meta-teacher $MT_t$'s teaching performance.
	Therefore, $MT_v$ greatly affects the evaluation of $MT_t$'s teaching performance.
	An improper $MT_v$ may destroy the premise defined in Section~\ref{Sec:outer-up1} and further causes $MT_v$ to mis-evaluate $MT_t$.
	For example, assume the worst situation, where the outputs of $MT_v$ are opposite to those of $MT_t$. The detector's validation loss evaluated based on $MT_v$ will incorrectly assess $MT_t$'s teaching ability.
	To avoid this situation, we have manually restricted the predictions of both $MT_t$ and $MT_v$ for live faces to zero-map $\otimes$ and constrained the pixel-wise predictions for spoof faces into the range (0, 2).
	Nevertheless, we still need to constrain the difference between the outputs of $MT_v$ and $MT_t$ for assessing $MT_t$'s teaching performance more precisely.
	To this end, in MT-FAS, we use a momentum update manner to update $MT_v$ in every iteration, which can be formulated as
	\begin{equation}
	\widehat{\omega} =  \gamma \cdot \widehat{\omega} + (1-\gamma) \cdot \omega,
	\label{eq:omega_update}
	\end{equation}
	where $\gamma$ is a hyper-parameter that is set to 0.999 by default.
	We will study the effect of $\gamma$ on the meta-teacher $MT_t$'s performance in Section~\ref{sec:gamma}.
	Another benefit of the momentum update manner of $MT_v$ is that in the training process, the optimization of $MT_t$ is probably rough and not smooth.
	Updating $MT_v$ with Eq.~\ref{eq:omega_update} smooths the updating, filters out the noise from $MT_t$, and further stabilizes the evaluation and training of $MT_t$.

	\subsubsection{Updating the detector}
	According to Section~\ref{Sec:inner-up}-\ref{Sec:outer-up2}, the higher-level learning only optimizes $\omega$ without updating $\theta$.
	However, ignoring updating $\theta$ will cause the representation gap between the detector and $MT_t$ to turn larger and larger with the progressively optimizing of $MT_t$.
	This may harm the evaluation and optimization of the meta-teacher.
	Therefore, for the detector to adapt to the updating $MT_t$, we periodically update $\theta$ by assigning it the weight $\theta^*(\omega)$ optimized in the lower-level learning.
	Lines 11 and 12 of Algorithm~\ref{algorithm:Meta-teacher} describe the periodic updating of the detector with the interval $T$.
	$T$ is set to 10 by default.
	Experiments described in Section~\ref{sec:update_detector} will show the importance of updating the detector.

	\subsubsection{Avoiding minor prediction}
	When minimizing $\mathcal{L}_{\Phi_v}(\theta^*(\omega); \widehat{\omega})$, we should avoid the possible meaningless optima where the meta-teacher outputs zero-map $\otimes$ for all faces. 
	This existing of the meaningless optima is caused by 1) we manually let $MT_t$ outputting zero-map $\otimes$ for all live faces; 2) we use momentum updating defined in Eq.~\ref{eq:omega_update} to update $MT_v$.
	Thus, $MT_t$ may learn to make its prediction map $2\sigma(g_{\omega}(x_k))$ be closer and closer to zero-map, since the smaller the pixel values of $2\sigma(g_{\omega}(x_k))$ are, the smaller the pixel-values of $MT_v$'s output $2\sigma(g_{\widehat{\omega}}(x_k))$, and finally the smaller the loss $\mathcal{L}_{\Phi_v}(\theta^*(\omega); \widehat{\omega})$.
	
	To avoid this meaningless optimization, we encourage $MT_t$ and $MT_v$ to output maps $2\sigma(g_{\omega}(x_k))$ and $2\sigma(g_{\widehat{\omega}}(x_k))$ containing appropriate large pixel values by adding another item $\beta\cdot\nabla_{\omega} \frac{\mu}{N_t} \sum_{k}^{N_t} mean(\sigma(g_{\omega}(x_k)))$ to Eq.~\ref{eq:label_optimize5}.
	Finally, $\omega$ is optimized with
	\begin{equation}
	\begin{aligned}
	\omega^{'} \! \! = \! \omega \! + \! \beta  \cdot  &[\alpha \! \cdot \! \nabla_{\omega} \nabla_{\theta} \mathcal{L}_{\Phi_t}(\theta; \omega)  \! \cdot \! \nabla_{\theta^*(\omega)} \mathcal{L}_{\Phi_v}(\theta^*(\omega); \widehat{\omega}) \\
	&+ \nabla_{\omega} \frac{\mu}{N_t} \sum_{k}^{N_t} mean(\sigma(g_{\omega}(x_k)))].  
	\end{aligned}
	\label{eq:label_optimize6}
	\end{equation}
	$mean(\sigma(g_{\omega}(x_k)))$ is the average value of all pixels in $g_{\omega}(x_k)$.
	$\mu$ with the default value of 0.001 is a hyper-parameter that controls how strongly we encourage $MT_t$ to output larger map.
	
	\begin{figure}[]
		\centering
		\includegraphics[width=0.8\columnwidth]{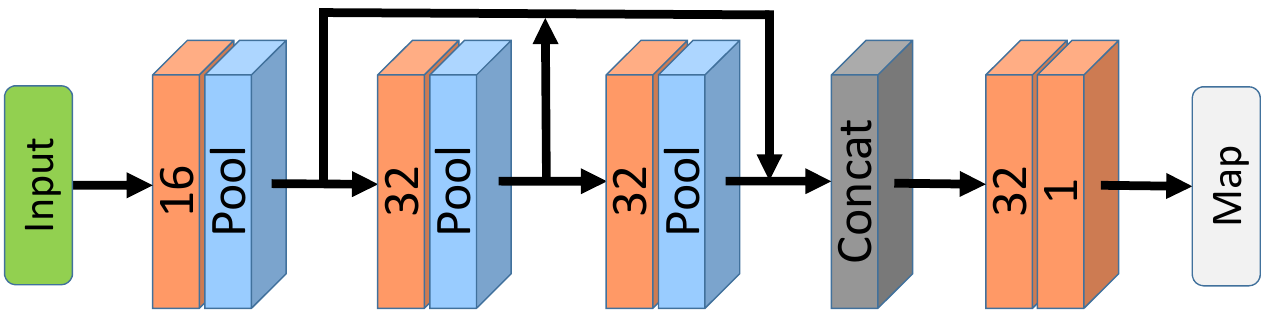}
		\caption{
			Network structure of FAS-DR-Light. 
			Each orange cube is the convolution layer and the number on it means the number of filters. The Pool layer is Max-pooling.
		}
		\label{fig:network}
	\end{figure}
	
	\subsubsection{Avoiding out-of-memory}
	The training of the meta-teacher $MT_t$ entails a high memory cost according to Eq.~\ref{eq:label_optimize5} because
	i) the losses $\mathcal{L}_{\Phi_t}$ and $\mathcal{L}_{\Phi_v}$ are calculated on all the faces from the training set $\Phi_t$ and validation set $\Phi_v$, respectively; and
	ii) second-order gradient is needed to optimize $MT_t$.
	We use the following two solutions to reduce memory consumption.
	First, we employ randomly sampled data batch as $\Phi_t$ and use another randomly sampled data batch as $\Phi_v$.
	Lines 4 and 5 in Algorithm\ref{algorithm:Meta-teacher} describe the data sampling process.
	According to the data sampling process, we can replace $N_t$ and $N_v$ in Eq.~\ref{eq:loss_MSE_train}, Eq.~\ref{eq:loss_MSE_val}, and Eq.~\ref{eq:label_optimize6} with $2M$ and $2N$, respectively.
	\textbf{Note that both the data batches $\Phi_t$ and $\Phi_v$ are sampled from the training set, and there are no overlapping examples between $\Phi_t$ and $\Phi_v$.}
	
	Second, we set all network structures of $MT_t$, $MT_v$, and the detector to a light version of FAS-DR\cite{qin2019learning} and name this version FAS-DR-Light.
	The network structure of FAS-DR-Light is shown in Fig.~\ref{fig:network}.
	We also show FAS-DR in Fig.~\ref{fig:FAS-DR-BC} for clarity.
	Note that our goal is to train a meta-teacher $MT_t$ to provide better pixel-wise supervision to train the detector.
	Thus, the detector here is a surrogate detector because it is only used to assist the meta-teacher's training.
	After training the meta-teacher $MT_t$, we use the trained $MT_t$ to supervise the existing state-of-the-art \cite{qin2019learning,DTN,shao2019regularized,yu2020searching2} detectors for better FAS performances.

	\section{Experiments}
	\label{sec:experiment}
	\subsection{Experimental Setup}
	\noindent \textbf{Performance Metrics.} \quad
	We use the following metrics in our experiment.
	1) Attack Presentation Classification Error Rate ($APCER$), {which denotes the ratio that spoof faces are misclassified into live faces.}
	2) Bona Fide Presentation Classification Error Rate ($BPCER$), {which denotes the ratio that live faces are misclassified into spoof faces.}
	3) Average Classification Error Rate ($ACER$)~\cite{ACER}, which evaluates the mean of $APCER$ and $BPCER$.
	4) Area Under Curve ($AUC$) {, which denotes the area under the Receiver Operating Characteristic (\emph{ROC}) curve}.
	5) Half Total Error Rate ($HTER$), { which denotes the mean of the False Acceptance Rate (\textit{FAR}) and False Rejection Rate (\textit{FRR})\cite{umphress1985identity}}.

\begin{figure*}[]
		\centering
		\includegraphics[width=0.99\textwidth]{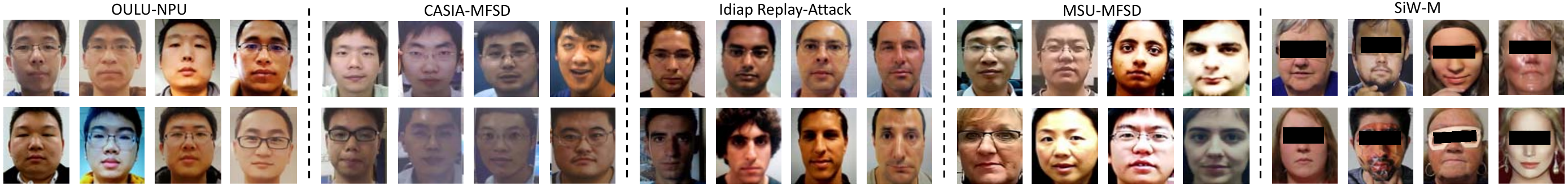}
		\caption{
			Examples from OULU-NPU\cite{Boulkenafet2017OULU}, CASIA-MFSD\cite{Zhang2012A}, Idiap Replay-Attack\cite{Chingovska2012On}, MSU-MFSD\cite{Wen2015Face}, and SiW-M\cite{DTN}.
			In each dataset, the faces in the leftmost column are live faces while all the other faces are spoof faces.
		}
		\label{fig:dataset}
	\end{figure*}

	\noindent \textbf{Experimental Datasets.} \quad
	We evaluate the developed MT-FAS on several popular FAS datasets, including OULU-NPU\cite{Boulkenafet2017OULU}, SiW-M\cite{DTN}, CASIA-MFSD\cite{Zhang2012A}, Idiap Replay-Attack\cite{Chingovska2012On}, and MSU-MFSD\cite{Wen2015Face}.
	We show some examples of these datasets in Fig.~\ref{fig:dataset} for a better understanding of live and spoof faces.
	
	OULU-NPU\cite{Boulkenafet2017OULU} is one of the most commonly used FAS dataset. 
	It contains several face capture conditions (six cameras and three sessions) and two kinds of printed spoof face and two kinds of replayed spoof face.
	Four protocols are used to evaluate PA detector's performance.
	Protocols 1, 2, and 3 evaluates the detector's performance on cross-camera, cross-session, cross-spoof-type scenarios, respectively.
	Protocol 4 is the more challenging because it evaluates the detector on the scenario of simultaneously cross-camera, cross-session, and cross-spoof-type.
	
	SiW-M\cite{DTN} is a recently proposed zero-shot FAS dataset.
	It contains 13 kinds of spoof attacks, such as print attack, 3D-Mask attack, Impersonation, Mannequin, and \emph{etc.}
	13 leave-one-attack-out sub-protocols are used to evaluate PA detectors' performance on novel spoof types.
	In each sub-protocol, the test set is formed with a part of live faces and one spoof attack, and the training set is formed with the other live faces and the other 12 spoof attacks.
	
	{CASIA-MFSD\cite{Zhang2012A} contains live and spoof faces captured with 50 genuine subjects.
	Three kinds of attack manners (warped photo-attack, cut photo attack, and video attack) are used to create spoof faces, and each facial image is recorded with three kinds of imaging qualities (low quality, normal quality, and high quality).
	Therefore, each subject has 3 kinds of live faces captured with different imaging qualities and has 3$\times$3=9 kinds of spoof faces captured with different attack manners and different imaging qualities.
	
	Idiap Replay-Attack\cite{Chingovska2012On} captures all live and spoof faces from 50 clients under two different lighting conditions.
	Five attack manners including four kinds of replayed faces and one kind of printed face are used to capture spoof faces.
	
	MSU-MFSD\cite{Wen2015Face} uses two different cameras to record all live and spoof faces from 35 genuine subjects.
	Three kinds of spoof faces are considered, including two kinds of replayed faces and one kind of printed face.
	Therefore, each subject has 2 kinds of live faces and has 2$\times$3=6 kinds of spoof faces captured with the two cameras.}
	
	Cross-domain FAS is a popular problem for practical FAS deployment.
	A domain-generalization benchmark\cite{shao2019multi} is commonly used to evaluate PA detector on this problem.
	This benchmark contains four datasets (OULU-NPU, CASIA-MFSD, Idiap Replay-Attack, and MSU-MFSD) and four cross-domain protocols.
	Each protocol uses one dataset as the testing domain while the other three datasets as the training domain.

	\noindent \textbf{Hyper-parameter Setup.} \quad
	We train $MT_t$ for 30,000 iterations (20,000 pretraining (Line 1 of Algorithm~\ref{algorithm:Meta-teacher}) + 10,000 bi-level-training iterations (Lines 4-14 of Algorithm~\ref{algorithm:Meta-teacher})).
	The two hyper-parameters $M$ and $N$ in Algorithm~\ref{algorithm:Meta-teacher} are set to 20 and 10, respectively.
	Both the learning rates $\beta$ and $\alpha$ are set to 0.001.
	On each mini training data batch $\Phi_t$, the meta-teacher $MT_t$ supervises the surrogate detector to do two gradient descent steps.
	$\mu$, $\gamma$, and $T$ are set to 0.001, 0.999, and 10, respectively.
	Both the network structures of $MT_t$, $MT_v$ is FAS-DR-Light illustrated in Fig.~\ref{fig:network}.
	FAS-DR-Light's output for each face is a single-channel pixel-wise map with 32$\times$32 resolution.
	All facial images used in this work are RGB images with 256$\times$256 resolution.
	After training the meta-teacher $MT_t$, we use Eq.~\ref{eq:normalized_supervision} as the final pixel-wise supervision to train PA detectors.
	\begin{equation}
	P_{\omega}(x_k) \! = \!
	\left\{ \! \! \! \!
	\begin{array}{lr}
	\qquad \qquad \quad \otimes,  \qquad \qquad \quad \qquad \  if \ x_k \in live   \\
	\frac{\sigma(g_{\omega}(x_k)) - min(\sigma(g_{\omega}(x_k)))}{max(\sigma(g_{\omega}(x_k))) - min(\sigma(g_{\omega}(x_k)))}, \;  if \ x_k \in spoof
	\end{array}
	\right.
	\label{eq:normalized_supervision}
	\end{equation}

	\noindent \textbf{Detector Nomenclature.}
	Our work aims to train a meta-teacher providing better pixel-wise supervision for PA detectors.
	To evaluate the trained meta-teacher, we use the meta-teacher to supervise the existing PA detectors' training.
	We call the detector that is supervised by the meta-teacher Detector(MT).
	For example, we call CDCN\cite{yu2020searching2} supervised by the meta-teacher CDCN(MT).
	For comparing the meta-teacher with existing handcrafted pixel-wise supervision, we also perform experiments in which we use existing handcrafted pixel-wise supervisions to train existing detectors.
	We name these trained detectors using the fashion of Detector(Label).
	For example, FAS-DR(Depth) denotes the detector FAS-DR\cite{qin2019learning} supervised by facial depth label.

	\subsection{Comparison with Handcrafted Labels}
	In this subsection, we verify the advantage of the proposed MT-FAS by comparing the performances of the detectors supervised by the meta-teacher with those of the counterpart detectors supervised by handcrafted labels.
	
	\begin{table}[t]
		\centering
		\caption{Experimental Results on OULU-NPU~\cite{Boulkenafet2017OULU}. 
		}
		\resizebox{0.49\textwidth}{!}{
			\begin{tabular}{|c|l|c|c|c|}
				\hline
				Protocol & Method & APCER(\%) & BPCER(\%) & ACER(\%) \\
				\hline
				\multirow{11}{*}{1}
				&GRADIANT ~\cite{boulkenafet2017competition}&1.3 &12.5 & 6.9 \\
				&STASN ~\cite{yang2019face} &1.2 &2.5 & 1.9 \\
				&Auxiliary ~\cite{Liu2018Learning} &1.6 &1.6 & 1.6 \\
				&FaceDs ~\cite{jourabloo2018face} &1.2 &1.7 & 1.5 \\
				&FAS-SGTD ~\cite{zezheng2020deep} &2.0 &0.0 & 1.0 \\
				&Disentangled ~\cite{zhang2020face} &1.7 &0.8 & 1.3 \\
				&DeepPixBiS ~\cite{george2019deep}&0.8 &0.0 & \textbf{0.4} \\
				&FAS-DR(Depth) & 0.7 & 2.3 & 1.5 \\
				&\textbf{FAS-DR(MT)} & 0.0 & 1.2 & 0.6 \\
				&CDCN(Depth)\cite{yu2020searching2} & 0.4 & 1.7 & 1.0 \\
				&\textbf{CDCN(MT)} & 0.0 & 0.8 & \textbf{0.4} \\
				\hline
				\multirow{11}{*}{2} 
				&DeepPixBiS ~\cite{george2019deep}&11.4 &0.6 & 6.0 \\
				&FaceDs ~\cite{jourabloo2018face}&4.2 &4.4 & 4.3 \\
				&Auxiliary ~\cite{Liu2018Learning}&2.7 &2.7 & 2.7 \\
				&GRADIANT ~\cite{boulkenafet2017competition}&3.1 &1.9 & 2.5 \\
				&STASN ~\cite{yang2019face}&4.2 &0.3 & 2.2 \\
				&FAS-SGTD ~\cite{zezheng2020deep} &2.5 &1.3 & 1.9 \\
				&Disentangled ~\cite{zhang2020face} &1.1 &3.6 & 2.4 \\
				&FAS-DR(Depth) &1.6  &3.4  &2.5 \\
				&\textbf{FAS-DR(MT)} & 0.9     & 2.7     &{1.8} \\
				&CDCN(Depth)\cite{yu2020searching2} &1.5  &1.4  &1.5 \\
				&\textbf{CDCN(MT)} &1.3  &1.4  &\textbf{1.4} \\
				\hline
				\multirow{11}{*}{3} 
				&DeepPixBiS ~\cite{george2019deep}&11.7$\pm$19.6 &10.6$\pm$14.1 & 11.1$\pm$9.4 \\
				&GRADIANT ~\cite{boulkenafet2017competition}&2.6$\pm$3.9 &5.0$\pm$5.3 &3.8$\pm$2.4 \\
				&FaceDs ~\cite{jourabloo2018face}&4.0$\pm$1.8 &3.8$\pm$1.2 &3.6$\pm$1.6 \\
				&Auxiliary ~\cite{Liu2018Learning}&2.7$\pm$1.3 &3.1$\pm$1.7 &{2.9}$\pm$1.5 \\
				&STASN ~\cite{yang2019face}&4.7$\pm$3.9 &0.9$\pm$1.2  &2.8$\pm$1.6 \\
				&FAS-SGTD ~\cite{zezheng2020deep}&3.2$\pm$2.0 &2.2$\pm$1.4 & 2.7$\pm$0.6 \\
				&Disentangled ~\cite{zhang2020face} &2.8$\pm$2.2 &1.7$\pm$2.6 & 2.2$\pm$2.2 \\
				&FAS-DR(Depth) &1.9$\pm$1.4 &5.8$\pm$7.5  &3.8$\pm$3.5 \\
				&\textbf{FAS-DR(MT)} &1.0$\pm$0.8 &3.8$\pm$4.1  &{2.4$\pm$2.1} \\
				&CDCN(Depth)\cite{yu2020searching2}      &2.4$\pm$1.3 &2.2$\pm$2.0  &2.3$\pm$1.4 \\
				&\textbf{CDCN(MT)}  &2.3$\pm$1.5 &1.9$\pm$1.8  &\textbf{2.1$\pm$1.7} \\
				\hline
				\multirow{11}{*}{4} 
				&DeepPixBiS ~\cite{george2019deep}&36.7$\pm$29.7 &13.3$\pm$14.1 & 25.0$\pm$12.7 \\
				&GRADIANT ~\cite{boulkenafet2017competition}&5.0$\pm$4.5 &15.0$\pm$7.1 &10.0$\pm$5.0 \\
				&Auxiliary ~\cite{Liu2018Learning}&9.3$\pm$5.6 &10.4$\pm$6.0 &9.5$\pm$6.0 \\
				&STASN ~\cite{yang2019face}&6.7$\pm$10.6 &8.3$\pm$8.4  &7.5$\pm$4.7 \\
				&FaceDs ~\cite{jourabloo2018face}&1.2$\pm$6.3 &6.1$\pm$5.1 &5.6$\pm$5.7 \\
				&FAS-SGTD ~\cite{zezheng2020deep}&6.7$\pm$7.5 &3.3$\pm$4.1 & 5.0$\pm$2.2 \\
				&Disentangled ~\cite{zhang2020face} &5.4$\pm$2.9 &3.3$\pm$6.0 & 4.4$\pm$3.0 \\
				&FAS-DR(Depth) &5.4$\pm$4.9 &8.2$\pm$7.8  &6.8$\pm$5.2 \\
				&\textbf{FAS-DR(MT)} &2.0$\pm$2.2 &6.6$\pm$5.7  &{4.3$\pm$4.0} \\
				&CDCN(Depth)\cite{yu2020searching2} &4.6$\pm$4.6 &9.2$\pm$8.0  &6.9$\pm$2.9 \\
				&\textbf{CDCN(MT)} &0.9$\pm$2.0 &6.4$\pm$4.9  &\textbf{3.7$\pm$2.9} \\
				\hline
			\end{tabular}
		}
		\label{tab:OULU}
	\end{table}

	\newcommand{\tabincell}[2]{\begin{tabular}{@{}#1@{}}#2\end{tabular}}
	\begin{table*}
		\centering
		\caption{Experimental Results on SiW-M~\cite{DTN} with leave-one-attack-out protocol.}
		\scalebox{0.7}{\begin{tabular}{c|c|c|c|c|c|c|c|c|c|c|c|c|c|c|c}
				\hline
				\multirow{2}{*}{Method} &\multirow{2}{*}{Metrics(\%)} &\multirow{2}{*}{Replay} &\multirow{2}{*}{Print} &\multicolumn{5}{c|}{Mask Attacks} &\multicolumn{3}{c|}{Makeup Attacks}&\multicolumn{3}{c|}{Partial Attacks} &\multirow{2}{*}{Average} \\
				\cline{5-15} &  &  &  & \tabincell{c}{Half} &\tabincell{c}{Silicone} &\tabincell{c}{Trans.} &\tabincell{c}{Paper}&\tabincell{c}{Manne.}&\tabincell{c}{Obfusc.}&\tabincell{c}{Imperson.}&\tabincell{c}{Cosmetic}&\tabincell{c}{Funny Eye} & \tabincell{c}{Paper Glasses} &\tabincell{c}{Partial Paper} & \\
				\hline
				\hline
				\multirow{4}{*}{SVM+LBP~\cite{Boulkenafet2017OULU}} 
				& APCER & 19.1 & 15.4 & 40.8 & 20.3 & 70.3 & 0.0 & 4.6 & 96.9 & 35.3 & 11.3 & 53.3 & 58.5 & 0.6 & 32.8$\pm$29.8 \\
				\cline{3-15}  & BPCER & 22.1 & 21.5 & 21.9 & 21.4 & 20.7 & 23.1 & 22.9 & 21.7 & 12.5 & 22.2 & 18.4 & 20.0 & 22.9 & 21.0$\pm$2.9 \\
				\cline{3-15}  & ACER & 20.6 & 18.4 & 31.3 & 21.4 & 45.5 & 11.6 & 13.8 & 59.3 & 23.9 & 16.7 & 35.9 & 39.2 & 11.7 & 26.9$\pm$14.5 \\
				\cline{3-15}  & EER & 20.8 & 18.6 & 36.3  & 21.4 & 37.2 & 7.5 & 14.1 & 51.2 & 19.8 & 16.1 & 34.4 & 33.0 & 7.9 & 24.5$\pm$12.9 \\
				\hline
				\hline
				\multirow{4}{*}{Auxiliary~\cite{Liu2018Learning}} & APCER & 23.7 & 7.3 & 27.7 & 18.2 & 97.8 & 8.3 & 16.2 & 100.0 & 18.0 & 16.3 & 91.8 & 72.2 & 0.4 & 38.3$\pm$37.4 \\
				\cline{3-15}  & BPCER & 10.1 & 6.5 & 10.9 & 11.6 & 6.2 & 7.8 & 9.3 & 11.6 & 9.3 & 7.1 & 6.2 & 8.8 & 10.3 & 8.9$\pm$2.0 \\
				\cline{3-15}  & ACER & 16.8 & 6.9 & 19.3 & 14.9 & 52.1 & 8.0 & 12.8 & 55.8 & 13.7 & 11.7 & 49.0 & 40.5 & 5.3 & 23.6$\pm$18.5 \\
				\cline{3-15}  & EER & 14.0 & {4.3} & 11.6  & 12.4 & 24.6 & 7.8 & 10.0 & 72.3 & 10.1 & \textbf{9.4} & 21.4 & 18.6 & 4.0 & 17.0$\pm$17.7 \\
				\hline
				\hline
				\multirow{4}{*}{CDCN++\cite{yu2020searching2}} & APCER & 9.2 & 6.0 & 4.2 & 7.4 & 18.2 & 0.0 & 5.0 & 39.1 & 0.0 & 14.0 & 23.3 & 14.3 & 0.0 & 10.8$\pm$11.2 \\
				\cline{3-15}  & BPCER & 12.4 & 8.5 & 14.0 & 13.2 & 19.4 & 7.0 & 6.2 & 45.0 & 1.6 & 14.0 & 24.8 & 20.9 & 3.9 & 14.6$\pm$11.4 \\
				\cline{3-15}  & ACER & 10.8 & 7.3 & {9.1} & 10.3 & 18.8 & 3.5 & 5.6 & 42.1 & 0.8 & 14.0 & 24.0 & 17.6 & 1.9 & 12.7$\pm$11.2 \\
				\cline{3-15}  & EER & 9.2 & 5.6 & \textbf{4.2}  & 11.1 & 19.3 & 5.9 & 5.0 & 43.5 & \textbf{0.0} & 14.0 & 23.3 & \textbf{14.3} & \textbf{0.0} & 11.9$\pm$11.8 \\
				\hline
				\hline				
				\multirow{4}{*}{BCN\cite{yu2020face}} & APCER & 12.4 & 5.2 & 8.3 & 9.7 & 13.6 & 0.0 & 2.5 & 30.4 & 0.0 & 12.0 & 22.6 & 15.9 & 1.2 & 10.3$\pm$9.1 \\
				\cline{3-15}  & BPCER & 13.2 & 6.2 & 13.1 & 10.8 & 16.3 & 3.9 & 2.3 & 34.1 & 1.6 & 13.9 & 23.2 & 17.1 & 2.3 & 12.2$\pm$9.4 \\
				\cline{3-15}  & ACER & 12.8 & 5.7 & 10.7 & 10.3 & 14.9 & \textbf{1.9} & \textbf{2.4} & \textbf{32.3} & 0.8 & 12.9 & 22.9 & \textbf{16.5} & 1.7 & 11.2$\pm$9.2 \\
				\cline{3-15}  & EER & 13.4 & 5.2 & 8.3  & 9.7 & 13.6 & 5.8 & \textbf{2.5} & \textbf{33.8} & \textbf{0.0} & 14.0 & 23.3 & 16.6 & 1.2 & 11.3$\pm$9.5 \\
				\hline
				\hline				
				\multirow{4}{*}{STDN\cite{liu2020disentangling}} & APCER & 1.6 & 0.0 & 0.5 & 7.2 & 9.7 & 0.5 & 0.0 & 96.1 & 0.0 & 21.8 & 14.4 & 6.5 & 0.0 & 12.2$\pm$26.1 \\
				\cline{3-15}  & BPCER & 14.0 & 14.6 & 13.6 & 18.6 & 18.1 & 8.1 & 13.4 & 10.3 & 9.2 & 17.2 & 27.0 & 35.5 & 11.2 & 16.2$\pm$7.6 \\
				\cline{3-15}  & ACER & 7.8 & 7.3 & \textbf{7.1} & 12.9 & 13.9 & 4.3 & 6.7 & 53.2 & 4.6 & 19.5 & 20.7 & 21.0 & 5.6 & 14.2$\pm$13.2 \\
				\cline{3-15}  & EER & \textbf{7.6} & \textbf{3.8} & 8.4  & 13.8 & 14.5 & 5.3 & {4.4} & {35.4} & \textbf{0.0} & 19.3 & 21.0 & 20.8 & 1.6 & 12.0$\pm$10.0 \\
				\hline
				\hline	
				\multirow{4}{*}{DTN\cite{DTN}} 
				& APCER & 1.0 & 0.0 & 0.7 & 24.5 & 58.6 & 0.5 & 3.8 & 73.2 & 13.2 & 12.4 & 17.0 & 17.0 & 0.2 & 17.1$\pm$23.3 \\
				\cline{3-15}  & BPCER & 18.6 & 11.9 & 29.3 & 12.8 & 13.4 & 8.5 & 23.0 & 11.5 & 9.6 & 16.0 & 21.5 & 22.6 & 16.8 & 16.6$\pm$6.2 \\
				\cline{3-15}  & ACER & {9.8} & 6.0 & 15.0 & 18.7 & 36.0 & 4.5 & 7.7 & 48.1 & 11.4 & 14.2 & 19.3 & 19.8 & 8.5 & 16.8$\pm$11.1 \\
				\cline{3-15}  & EER & 10.0 & 2.1 & 14.4 & 18.6 & 26.5 & 5.7 & 9.6 & 50.2 & 10.1 & 13.2 & 19.8 & 20.5 & 8.8 & 16.1$\pm$12.2 \\
				\hline
				\hline
				\multirow{4}{*}{\textbf{DTN(MT)}} 
				& APCER & 6.1 & 5.2 & 8.3 & 14.8 & 24.3 & 5.9 & 5.0 & 39.4 & 5.1 & 10.0 & 17.1 & 19.8 & 1.1 & 12.5$\pm$10.2 \\
				\cline{3-15}  
				& BPCER & 10.9 & 10.1 & 17.8 & 18.6 & 16.9 & 0.0 & 6.2 & 28.9 & 2.4 & 14.7 & 20.9 & 21.7 & 6.7 & 13.5$\pm$8.0 \\
				\cline{3-15}  
				& ACER & {9.5} & 7.6 & 13.1 & 16.7 & 20.6 & 2.9 & 5.6 & 34.2 & 3.8 & 12.4 & \textbf{19.0} & 20.8 & 3.9 & 13.1$\pm$8.7 \\
				\cline{3-15}  
				& EER & 9.1 & 7.8 & 14.5 & 14.1 & 18.7 & 3.6 & 6.9 & 35.2 & 3.2 & 11.3 & 18.1 & 17.9 & 3.5 & 12.6$\pm$8.5 \\
				
				\hline
				\hline	
				\multirow{4}{*}{FAS-DR(Depth)}
				& APCER  & 9.3 & 6.0 & 15.9 & 11.9 & 19.6 & 7.2 & 8.9 & 38.2 & 2.9 & 14.9 & 20.0 & 20.1 & 1.1 & 13.5$\pm$9.4 \\
				\cline{3-15}  & BPCER  & 6.3 & 5.8 & 10.9 & 11.6 & 15.2 & 3.5 & 6.0 & 39.7 & 1.8 & 10.4 & 19.3 & 16.7 & 3.8 & 11.6$\pm$9.6 \\
				\cline{3-15}  & ACER   & 7.8 & 5.9 & 13.4 & 11.7 & 17.4 & 5.4 & 7.4 & 39.0 & 2.3 & 12.6 & 19.6 & 18.4 & 2.4 & 12.6$\pm$9.5\\
				\cline{3-15}  & EER    & 8.0 & 4.9 & 10.8 & 10.2 & 14.3 & 3.9 & 8.6 & 45.8 & 1.0 & 13.3 & \textbf{16.1} & 15.6 & 1.2 & 11.8$\pm$11.0 \\
				\hline
				\hline
				\multirow{4}{*}{\textbf{FAS-DR(MT)} }
				& APCER  & 4.1 & 4.3 & 6.5 & 3.7 & 9.2 & 5.9 & 5.0 & 36.4 & 0.0 & 10.0 & 20.3 & 17.5 & 0.0 & 9.5$\pm$9.7 \\
				\cline{3-15}  
				& BPCER  & 8.5 & 5.4 & 12.0 & 10.9 & 14.7 & 0.8 & 1.6 & 42.6 & 0.4 & 10.9  & 21.7 & 19.4 & 2.2 & 11.6$\pm$11.1 \\
				\cline{3-15}  
				& ACER   & \textbf{6.3} & \textbf{4.9} & 9.3 & \textbf{7.3} & \textbf{12.0} & 3.3 & 3.3 & 39.5 & \textbf{0.2} & \textbf{10.4} & 21.0 & 18.4 & \textbf{1.1} & \textbf{10.5$\pm$10.3} \\
				\cline{3-15}  
				& EER    & {7.8} & 4.4 & 11.2 & \textbf{5.8} & \textbf{11.2} & \textbf{2.8} & 2.7 & 38.9 & 0.2 & 10.1  & 20.5 & 18.9 & 1.3 & \textbf{10.4$\pm$10.2} \\
				\hline
				\hline
			\end{tabular}
		}
		\label{tab:SiW-M}
	\end{table*}

	\subsubsection{Experiment on OULU-NPU}
	\label{sec:exp_oulu}
\begin{figure}[]
		\centering
		\includegraphics[width=0.99\columnwidth]{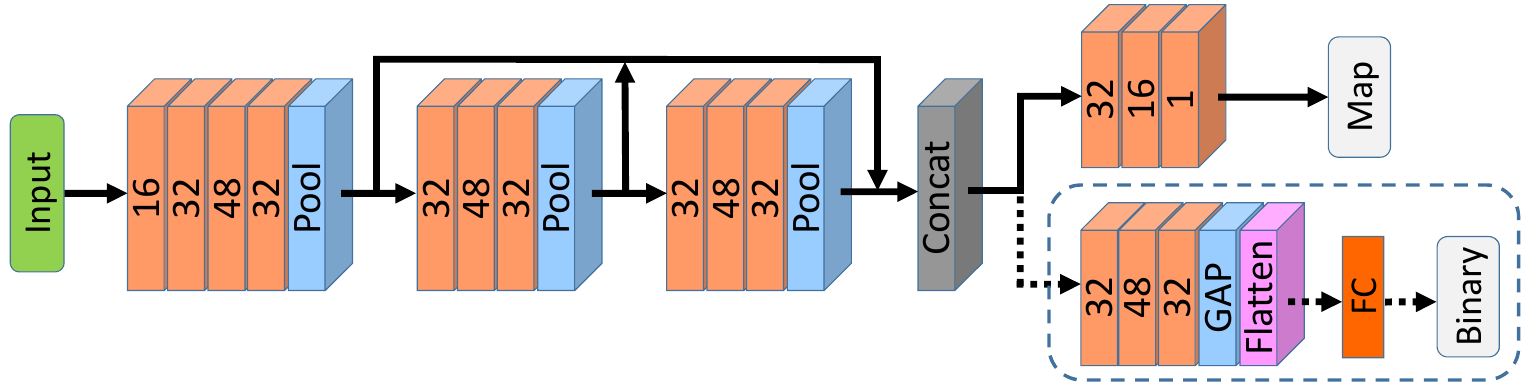}
		\caption{
			Network structure of FAS-DR and FAS-DR-BC. 
			FAS-DR is the network without the layers in the dashed box while FAS-DR-BC contains all layers in this figure.
			Each orange cube is the convolution layer and the number on it means the number of filters. The convolution layers in the dashed box use 2x2 stride while the other convolution layers use 1x1 stride.
			The Pool layer is Max-pooling with 2x2 stride and GAP is the global average pooling.
		}
		\label{fig:FAS-DR-BC}
	\end{figure}

	Here, we evaluate MT-FAS on OULU-NPU\cite{Boulkenafet2017OULU}.
	First, we train the meta-teacher $MT_t$ on protocol 1.
	Second, on all protocols, we use the trained $MT_t$ to supervise the learning of two existing PA detectors (FAS-DR and {CDCN\cite{yu2020searching2}}).
	The detailed network structure of FAS-DR\cite{qin2019learning} is shown in Fig.~\ref{fig:FAS-DR-BC}.
	{FAS-DR uses four convolution neural network-based blocks to regress the input face as a pixel-wise map.
	The bottom three blocks extract low-level, middle-level, and high-level image features.
	The features are then concatenated and fed into the top block to regress the pixel-wise map.
	CDCN processes the input face in a similar way.
	The most difference between FAS-DR and CDCN is that FAS-DR uses vanilla convolution layers to process images while CDCN uses central difference-based convolution layers.
	}
	
	{
	We denote the two trained detectors as FAS-DR(MT) and CDCN(MT) and report their performances in Table \ref{tab:OULU}.
	Both of them perform well on all protocols, especially CDCN(MT).
	Compared with other existing PA detectors, CDCN(MT) achieves the best performances on all protocols.
	For instance, CDCN(MT) decreases the $ACER$ by approximately 16\% on protocol 4.}
	
	{
	Note that, since the official CDCN is trained with facial depth supervision, we denote it as CDCN(Depth) for clarity.
	Moreover, we use facial depth label to train another detector FAS-DR(depth).
	FAS-DR(depth) and CDCN(Depth) can be treated as baselines for FAS-DR(MT) and CDCN(MT), respectively.
	The comparison between FAS-DR(depth) and FAS-DR(MT), and the comparison between CDCN(depth) and CDCN(MT) can verify the advantage of meta-teacher over the handcrafted facial depth label in supervising the two PA detectors.
}
	The experimental results reported in Table \ref{tab:OULU} show that on protocols 1-4, CDCN(MT) and FAS-DR(MT) achieve lower $ACER$ than CDCN(Depth) and FAS-DR(depth), respectively.
	For instance, compared with FAS-DR(depth), FAS-DR(MT) decreases $ACER$ by approximately 60\% on protocol 1.
	The comparisons demonstrate that compared with handcrafted facial depth label, the meta-teacher supervises the two detectors (FAS-DR and CDCN) more accurately.

	\begin{table*}[]
	\centering
	\caption{Experimental Results on the Domain-generalization Benchmark.
		The Metrics Used in This Experiment are $HTER$ and $AUC$.
	}
	\resizebox{0.85\textwidth}{!}{
		\begin{tabular}{|c|c|c|c|c|c|c|c|c|}
			\hline
			\multirow{2}{*}{\textbf{Method}} &\multicolumn{2}{c|}{\textbf{O\&C\&I to M}} &\multicolumn{2}{c|}{\textbf{O\&M\&I to C}}&\multicolumn{2}{c|}{\textbf{O\&C\&M to I}} &\multicolumn{2}{c|}{\textbf{I\&C\&M to O}} \\
			\cline{2-9} &{HTER(\%)} &{AUC(\%)} &{HTER(\%)} &{AUC(\%)}&{HTER(\%)}&{AUC(\%)}&{HTER(\%)}&{AUC(\%)} \\
			\hline
			MS\_LBP ~\cite{Maatta2011Face} 
			& 29.76 & 78.50 & 54.28 & 44.98 & 50.30 & 51.64 & 50.29 & 49.31 \\
			\hline
			CNN ~\cite{Yang2014Learn} 
			& 29.25 & 82.87 & 34.88 & 71.95 & 34.47 & 65.88 & 29.61 & 77.54 \\
			\hline
			IDA ~\cite{MSU} 
			& 66.67 & 27.86 & 55.17 & 39.05 & 28.35 & 78.25 & 54.20 & 44.59 \\
			\hline
			LBPTOP ~\cite{de2014face} 
			& 36.90 & 70.80 & 42.60 & 61.05 & 49.45 & 49.54 & 53.15 & 44.09 \\
			\hline
			Color Texture ~\cite{Boulkenafet2017Face} 
			& 28.09 & 78.47 & 30.58 & 76.89 & 40.40 & 62.78 & 63.59 & 32.71 \\
			\hline
			Auxiliary(Depth only) ~\cite{Liu2018Learning} 
			& 22.72 & 85.88 & 33.52 & 73.15 & 29.14 & 71.69 & 30.17 & 66.61 \\
			\hline
			MMD-AAE ~\cite{li2018domain} 
			& 27.08 & 83.19 & 44.59 & 58.29 & 31.58 & 75.18 & 40.98 & 63.08 \\
			\hline
			MADDG ~\cite{shao2019multi} 
			& 17.69 & 88.06 & 24.5 & 84.51 & 22.19 & 84.99 & 27.98 & 80.02 \\
			\hline
			DR\_MD ~\cite{wang2020cross} 
			& 17.02 & 90.10 & 19.68 & 87.43 & 20.87 & 86.72 & 25.02 & 81.47 \\
			\hline
			MA-Net ~\cite{liu2021face
			} 
			& 20.80 & / & 25.60 & / & 24.70 & / & 26.30 & / \\
			\hline
			RFMetaFAS ~\cite{shao2019regularized} 
			& 13.89 & 93.98 & 20.27 & 88.16 & 17.30 & 90.48 & 16.45 & 91.16 \\
			\hline
			RFMetaFAS*  
			& 11.90 & 94.65 & 24.76 & 84.29 & 18.89 & 88.71 & 21.83 & 85.56 \\
			\hline
			RFMetaFAS(MT)*  
			& 12.31 & \textbf{94.89} & 22.91 & 85.63 & 12.77 & 94.02 & 18.16 & 89.40 \\
			\hline
			FAS-DR-BC(Depth)
			& 13.81 & 91.61 & 19.67 & 89.36 & 19.14 & 87.85 & 19.56 & 88.28\\
			\hline
			\textbf{FAS-DR-BC(MT)} 
			& \textbf{11.67} & 93.09 & \textbf{18.44} & \textbf{89.67} & \textbf{11.93} & \textbf{94.95} & \textbf{16.23} & \textbf{91.18}\\
			\hline
	\end{tabular}}
	\label{tab:Domain generation}
\end{table*}

	\subsubsection{Experiment on SiW-M}
	\label{sec:exp_SIW}
	On each protocol of SiW-M\cite{DTN}, we first train the meta-teacher $MT_t$ and then use the trained meta-teacher to train the FAS-DR detector.
	We report the experimental results in Table \ref{tab:SiW-M}.
	FAS-DR(MT) outperforms the other state-of-the-art methods by a large margin. One highlight is that compared with DTN\cite{DTN}, FAS-DR(MT) decreases $ACER$ by approximately 37.5\%.

	To compare the supervision of the meta-teacher $MT_t$ with that of facial depth and pixel-wise binary mask labels, we further train the detector FAS-DR with facial depth label and train the detector DTN with the meta-teacher.
	The corresponding experimental results are shown in Table \ref{tab:SiW-M}.
	Compared with FAS-DR(Depth), FAS-DR(MT) decreases the average $ACER$ by approximately 17\%, and compared with DTN, DTN(MT) decreases the average $ACER$ by approximately 22\%.
	As the official DTN is trained using pixel-wise binary mask label (shown in Fig.~\ref{fig:fig1}), the comparison between the official DTN and DTN(MT) vividly reveals the advantage of the trained meta-teacher $MT_t$ over handcrafted pixel-wise binary mask label.
	The comparison between the official FAS-DR(Depth) and FAS-DR(MT) further reveals the advantage of the meta-teacher over facial depth supervision.
	
	{The possible underlining reasons why the meta-teacher outperforms handcrafted pixel-wise regression labels in supervising PA detectors are 1) the handcrafted labels are not the most suitable labels for the FAS problem; and 2) the meta-teacher is task-oriented trained to produce more suitable pixel-wise labels for FAS.}

	\subsubsection{Experiment on domain-generalization benchmark}
	\label{sec:exp_domain}
	In the domain-generalization benchmark\cite{shao2019multi}, four protocols are utilized to evaluate the PA detector's domain-generalization performance.
	In this experiment, we revise the FAS-DR network by adding a binary classification branch to it.
	We name the revised network FAS-DR-BC and illustrate its structure in Fig.~\ref{fig:FAS-DR-BC}.
	When testing, the prediction of FAS-DR-BC is the average score of the pixel-wise regression map and binary classification branches.
	Specifically, $\hat{y} = 0.5*mean(\hat{y}_{map}) + 0.5*\hat{y}_{binary}$, where $\hat{y}_{map}$ is the map regressed by the regression branch and $\hat{y}_{binary}$ is the score predicted by the classification branch.
	The score $\hat{y}_{binary}$ denotes the probability that the input face is a spoof face.
	
	On each protocol, we first train the meta-teacher on the corresponding training set and then use the meta-teacher to supervise FAS-DR-BC's regression output.
	We further assess the meta-teacher by using it to supervise the recent state-of-the-art detector RFMetaFAS\cite{shao2019regularized}.
	The official RFMetaFAS is trained using both facial depth label and binary classification label.
	In this experiment, we replace the facial depth supervision with the trained meta-teacher and keep all the other settings the same as those in the official RFMetaFAS.

	The experimental results are shown in Table \ref{tab:Domain generation}.
	O, C, I, and M denote OULU-NPU\cite{Boulkenafet2017OULU}, CASIA-MFSD\cite{Zhang2012A}, Idiap Replay-Attack\cite{Chingovska2012On}, and MSU-MFSD\cite{Wen2015Face}, respectively.
	``I\&C\&M to O" denotes the protocol where the PA detector is trained on I, C, and M, and tested on O; the same explanation holds for the other protocols.
	
	In this experiment, FAS-DR-BC(MT) achieves state-of-the-art performances in most cases.
	Compared with FAS-DR-BC(Depth), FAS-DR-BC(MT) decreases $HTER$ by approximately 15\%, 6\%, 38\%, and 17\%, on the four protocols.
	Besides, RFMetaFAS(MT)* also outperforms RFMetaFAS*.
	Note that RFMetaFAS* is our reimplementation of RFMetaFAS using the published official code with carefully tuned hyper-parameters.
	
	This experiment demonstrates 1) the meta-teacher's superiority over facial depth supervision when training PA detectors, 2) the generalization of the MT-FAS method on the domain-generalization benchmark, {and 3) the meta-teacher's generalizability on supervising different PA detectors.}

	\subsection{Comparison with existing teacher-student methods}
	In this subsection, we evaluate whether the proposed MT-FAS outperforms existing teacher-student methods in improving PA detectors' learning.
	The compared teacher-student methods include Distill\cite{hinton2015distilling}, BANs\cite{furlanello2018born}, and Fitnet\cite{romero2015fitnets}. 
	
	We implement the compared teachers on FAS, and use the pixel-wise binary label to train them regressing live faces as zero-maps and regressing spoof faces as one-maps.
	For the Distill and Fitnet teachers, we set their network to be deeper than FAS-DR by repeating each convolution layer containing 48 filters for 2 times in FAS-DR.
	For the BANs teacher, we set its network to be the same as FAS-DR.
	As our meta-teacher uses a shallow backbone FAS-DR-Light, another light version teacher Distill-Light which also uses the FAS-DR-Light backbone is trained as the baseline of our meta-teacher.
	
	For all the compared teachers, we train them on the training set for 10 epochs with the learning rate of 0.001.
	Adam is selected as the optimizer.
	After training the teachers, we apply them to supervise the FAS-DR detector's training.
	Note that to ensure a fair comparison between the proposed meta-teacher and the compared teachers, we modify the compared teachers' output with Eq.~\ref{eq:teachers} when using them to supervise the FAS-DR detector.
	In Eq.~\ref{eq:teachers}, $x$ is the input face image, and $\otimes$ is zero-map.
	$\omega$ is the teacher's weight and $g_\omega(x)$ is the teacher's output for spoof faces.
	
	\begin{equation}
	T_{\omega}(x)=
	\left\{
	\begin{array}{lr}
	\quad \otimes, \qquad \qquad \qquad \qquad \, if \ x \in live,   \\
	\frac{g_{\omega}(x) - min(g_{\omega}(x))}{max(g_{\omega}(x)) - min(g_{\omega}(x))}, \;  if \ x \in spoof
	\end{array}
	\right.
	\label{eq:teachers}
	\end{equation}
	
	With this modification, these teachers output pixel-wise predictions for spoof faces and output zero-maps for all live faces, which is consistent with the predictions of the proposed meta-teacher.
	We denote the trained detectors supervised by these teachers with the nomenclature fashion of Detector(Teacher).
	For example, FAS-DR(BANs) denotes the FAS-DR detector supervised by the BANs teacher.
	
	{OKDDip\cite{Chen_Mei_Wang_Feng_Chen_2020}, a teacher-free online knowledge-distillation method, is also considered as a baseline to MT-FAS. 
	OKDDip simultaneously uses $m$ (3 in our re-implementation) students where $m-1$ are auxiliary peers and one is group-leader to do knowledge distillation.
	It trains each auxiliary peer in the first-level distillation and uses the ensemble of auxiliary peers together with ground-truth to supervise the group-leader in the second-level distillation. 
	We treat the ensemble of auxiliary peers as another `teacher' because it supervises the group-leader.
	}
	
	\begin{table}[]
		\centering
		\caption{Comparison Between the Meta-teacher and Other Teachers on Protocol 4 of OULU-NPU~\cite{Boulkenafet2017OULU}. 
			Time$_T$ denotes the total training time (hours) of the teacher.
		}
		{\scriptsize 
			\begin{tabular}{|l|c|c|c|c|}
				\hline
				Method & ACER(\%) & {FLOPs$_{T}$} & {Time$_{T}$} & {FLOPs$_{S}$} \\
				\hline
				FAS-DR(Depth) &6.8$\pm$5.2 & / & / &{280G}\\
				\hline
				FAS-DR(Distill)\cite{hinton2015distilling}  &5.7$\pm$3.6 & {453G} & {13.6H} & {431G}\\
				\hline
				FAS-DR(BANs)\cite{furlanello2018born}  &6.1$\pm$5.4 & {280G} & {9.0H} & {379G} \\
				\hline
				FAS-DR(Fitnet)\cite{romero2015fitnets}  &5.5$\pm$3.9 & {453G} & {13.6H} & {436G} \\
				\hline
				FAS-DR(Distill\_Light)  &7.5$\pm$4.4 & {5.3G} & {1.5H} &{ 283G} \\
				\hline
				{FAS-DR(OKDDip)\cite{Chen_Mei_Wang_Feng_Chen_2020}}  &{5.4$\pm$3.6} & / & / &{862G}\\
				\hline
				\textbf{FAS-DR(MT)} &\textbf{4.3$\pm$4.0} & {781G} & {8.9H} & {283G}\\
				\hline
			\end{tabular}
		}
		\label{tab:OULU4}
	\end{table}

	\subsubsection{Experiment on OULU-NPU}
	\label{sec:exp2_oulu}
	As protocol 4 is the most challenging protocol in OULU-NPU, we implement the compared teachers on protocol 4.
	{Table \ref{tab:OULU4} reports the performances of FAS-DR trained using the compared teachers.
	Note that the network of all the students in OKDDip is set to FAS-DR and FAS-DR(OKDDip) denotes the trained group-leader.}
	The experimental results show that meta-teacher outperforms the compared teachers in supervising FAS-DR's learning.
	Compared with the other teachers, the meta-teacher decreases FAS-DR's $ACER$ by at least 20\%.

	{
	We also compare the training costs of all teachers in Table \ref{tab:OULU4}.
	FLOPs$_T$ and FLOPs$_S$ denote the training cost of the teacher model and the student model, respectively, in each training iteration.
	Time$_T$ denotes the total training time (hours) of the teacher.
	Note that all teachers are trained on one NVIDIA Tesla P40 GPU.
	FAS-DR(Depth) is also listed in Table \ref{tab:OULU4} as a baseline.
	As FAS-DR(Depth) is the detector FAS-DR trained using facial depth label without using teacher model, both its FLOPs$_T$ and Time$_T$ are zero (denoted as $/$).
	Both FLOPSs$_T$ and Time$_T$ of OKDDip\cite{Chen_Mei_Wang_Feng_Chen_2020} are zero too because OKDDip is an online knowledge distillation method that simultaneously trains all models without separately training the teacher and student.
	Therefore all FLOPs of OKDDip in each training iteration are summed into FLOPs$_S$.
}

	{
	Table \ref{tab:OULU4} shows that compared with the other teachers, the proposed MT-FAS costs more FLOPs to train the meta-teacher in each bi-level training iteration.
	The reason is that MT-FAS needs to calculate the second-order gradient shown in Eq.~\ref{eq:label_optimize6} to optimize the meta-teacher.
	But overall, compared with the other teachers, the meta-teacher does not cost more training time.
	This is because the meta-teacher needs fewer training iterations than the other methods to converge.
	The compared teachers cost about 40,000 iterations to converge while the meta-teacher needs fewer iterations (20,000 pretraining iterations (Line 1 of Algorithm~\ref{algorithm:Meta-teacher}) + 10,000 bi-level optimization iterations (Lines 4-14 of Algorithm~\ref{algorithm:Meta-teacher})).
}

	{
	Table \ref{tab:OULU4} does not show the training time of each detector (student) because we use the same training iterations to train all detectors.
	In other words, the training cost of each detector is mainly reflected by FLOPs$_S$.
	We can see that the training of FAS-DR(MT) costs fewer FLOPs than most of the other detectors including FAS-DR(Distill), FAS-DR(BANs), and \emph{etc.}
	The reason is that benefiting from the light-weight network FAS-DR-Light, the meta-teacher costs fewer computation resources than most of the other compared teachers in inference.
	}

		\begin{table*}[t]
		\centering
		\caption{Comparison Between the Meta-teacher and Other Teachers on the Domain-generalization Benchmark.
		}
		{\footnotesize 
			\begin{tabular}{|l|c|c|c|c|c|c|c|c|}
				\hline
				\multirow{2}{*}{\textbf{Method}} &\multicolumn{2}{c|}{\textbf{O\&C\&I to M}} &\multicolumn{2}{c|}{\textbf{O\&M\&I to C}}&\multicolumn{2}{c|}{\textbf{O\&C\&M to I}} &\multicolumn{2}{c|}{\textbf{I\&C\&M to O}} \\
				\cline{2-9} &{HTER(\%)} &{AUC(\%)} &{HTER(\%)} &{AUC(\%)}&{HTER(\%)}&{AUC(\%)}&{HTER(\%)}&{AUC(\%)} \\
				\hline
				FAS-DR-BC(Depth)
				& 13.81 & 91.61 & 19.67 & 89.36 & 19.14 & 87.85 & 19.56 & 88.28\\
				\hline
				FAS-DR-BC(Distill)\cite{hinton2015distilling}
				& 15.24 & 90.46 & 23.22 & 86.71 & 23.86 & 83.65 & 17.60 & 89.51\\
				\hline
				FAS-DR-BC(BANs)\cite{furlanello2018born}
				& \textbf{11.31} & \textbf{93.57} & 19.12 & 88.43 & 24.31 & 78.29 & 18.53 & 89.17\\
				\hline
				FAS-DR-BC(Fitnet)\cite{romero2015fitnets}
				& 12.53 & 91.79 & 18.65 & 89.58 & 21.14& 85.17 & 17.19 & 90.25\\
				\hline
				FAS-DR-BC(Distill\_Light)
				& 11.90 & 92.86 & 20.03 & 86.48 & 25.65& 79.10& 18.30 & 88.76\\
				\hline
				{FAS-DR-BC(OKDDip)\cite{Chen_Mei_Wang_Feng_Chen_2020}}
				& {12.25} & {92.36} & {19.81} & {88.52} & {19.75}& {86.29}& {17.22} & {90.06}\\
				\hline
				\textbf{FAS-DR-BC(MT)} 
				& 11.67 & 93.09 & \textbf{18.44} & \textbf{89.67} & \textbf{11.93} & \textbf{94.95} & \textbf{16.23} & \textbf{91.18}\\
				\hline
		\end{tabular}}
		\label{tab:Domain generation2}
	\end{table*}

	\subsubsection{Experiment on domain-generalization benchmark}
	\label{sec:exp2_domain}
	On the domain-generalization benchmark, we utilize the compared teachers to supervise the detector FAS-DR-BC's regression prediction.
	{Note that the teacher of OKDDip means the ensemble of auxiliary peers.
	The network of each auxiliary peer in OKDDip is set to FAS-DR and the network of the group-leader is set to FAS-DR-BC in this experiment.
	We use the ensemble of auxiliary peers together with the pixel-wise binary label to supervise the group-leader's regression prediction in the second-level distillation of OKDDip.}
	Table \ref{tab:Domain generation2} reports the corresponding experimental results of FAS-DR-BC supervised by these teachers.
	Obviously, FAS-DR-BC(MT) outperforms the other trained FAS-DR-BC counterparts, which verifies the advantage of the proposed MT-FAS over the teacher-student methods and OKDDip.

	All the aforementioned experiments validate that the proposed meta-teacher outperforms not only the widely employed human-designed labels but also existing teachers, in supervising PA detectors.
	{The possible reason for these experimental results is that existing teachers are trained to match the training data but not to improve their teaching ability. In contrast, the proposed meta-teacher is trained to learn how to supervise the student to perform better.}

	\subsection{Ablation Study}
	In this subsection, we evaluate how crucial components or settings affect the meta-teacher's performance.
	All ablation experiments are conducted with FAS-DR on protocol 1 of OULU-NPU.

	\subsubsection{Effect of pretraining}
	\label{sec:pretrain}
	In our implementation of MT-FAS, before optimizing the meta-teacher $MT_t$, we initialize $MT_t$, $MT_v$, and the surrogate detector by pretraining them on the training set.
	In this experiment, we optimize $MT_t$ from scratch without pretraining.
	We denote the meta-teacher trained without pretraining as MT\_w/o\_pre and denote the trained FAS-DR detector supervised by MT\_w/o\_pre as FAS-DR(MT\_w/o\_pre).
	The corresponding experimental result is shown in Table \ref{tab:Ablation}.
	Without pretraining, the detector FAS-DR's $ACER$ rises significantly from 0.6\% to 6.3\%, which reveals the importance of pretraining for the meta-teacher.
	
	\subsubsection{Indispensability of $MT_v$}
	\label{sec:exp_MT_v}
	When training the meta-teacher $MT_t$, as shown in Eq.~\ref{eq:loss_MSE_val}, we use another $MT_v$ to evaluate $MT_t$'s teaching quality.
	In this ablation experiment, we verify whether $MT_v$ is indispensable to evaluate $MT_t$.
	In other words, if we use $MT_t$ to calculate $\mathcal{L}_{\Phi_v}$ (realized by copying $\omega$ to $\widehat{\omega}$ at every training iteration), then can we still stably optimize $MT_t$ with Eq.~\ref{eq:label_optimize6}?
	
	We denote the meta-learner trained without $MT_v$ as MT\_w/o\_$MT_v$ and denote the trained detector supervised by MT\_w/o\_$MT_v$ as FAS-DR(MT\_w/o\_$MT_v$).
	The experimental result shown in Table \ref{tab:Ablation} apparently indicates that $MT_v$ is indispensable for evaluating $MT_t$'s teaching quality.
	Without $MT_v$, the FAS-DR detector's $ACER$ greatly deteriorates from 0.6\% to 7.1\%.
	
	\begin{table}[]
		\centering
		\caption{Ablation Experimental Results on Protocol 1 of OULU-NPU~\cite{Boulkenafet2017OULU}. 
		}
		{\footnotesize 
			\begin{tabular}{|l|c|c|c|}
				\hline
				Method & APCER(\%) & BPCER(\%) & ACER(\%) \\
				\hline
				FAS-DR(MT\_w/o\_$MT_v$) &0.4 &13.8 & 7.1 \\
				\hline
				FAS-DR(MT\_w/o\_pre) &1.0 & 11.5 & 6.3\\
				\hline
				FAS-DR(MT\_w/o\_adapt) & 2.8 & 5.6 & 4.2 \\
				\hline
				FAS-DR(Depth) & 0.7 & 2.3 & 1.5 \\
				\hline
				FAS-DR(MT\_resnet) &0.2 & 1.4 & 0.8\\
				\hline
				{{FAS-DR(MT)$_{64}$}} & {0.2} & {1.2} & {{0.7}} \\
				\hline
				{{FAS-DR(MT)$_{32}$}} & {0.4}& {0.8} & {\textbf{0.6}} \\
				\hline
				\textbf{FAS-DR(MT)} & 0.0 & 1.2 & {\textbf{0.6}} \\
				\hline
			\end{tabular}
		}
		\label{tab:Ablation}
	\end{table}

	\subsubsection{Momentum update hyper-parameter $\gamma$}
	\label{sec:gamma}
	In this ablation experiment, we aim to verify how $MT_v$'s momentum update hyper-parameter $\gamma$ affects the trained meta-teacher $MT_t$.
	$\gamma$ is set to 0.999 by default in this work.
	Here, we set $\gamma$ to other values of 0.9, 0.99, 0.995, 0.9995, 0.9999, and 1.0.
	According to Eq.~\ref{eq:omega_update}, the larger $\gamma$ is, the slower the update of $MT_v$.
	When $\gamma=1.0$, $MT_v$ will be frozen in the optimization procedure of $MT_t$.
	$\gamma=0.9$, $0.99$, and $0.995$ will update $MT_v$ faster than $\gamma=0.999$.

\begin{figure}[]
		\centering
		\includegraphics[width=0.9\columnwidth]{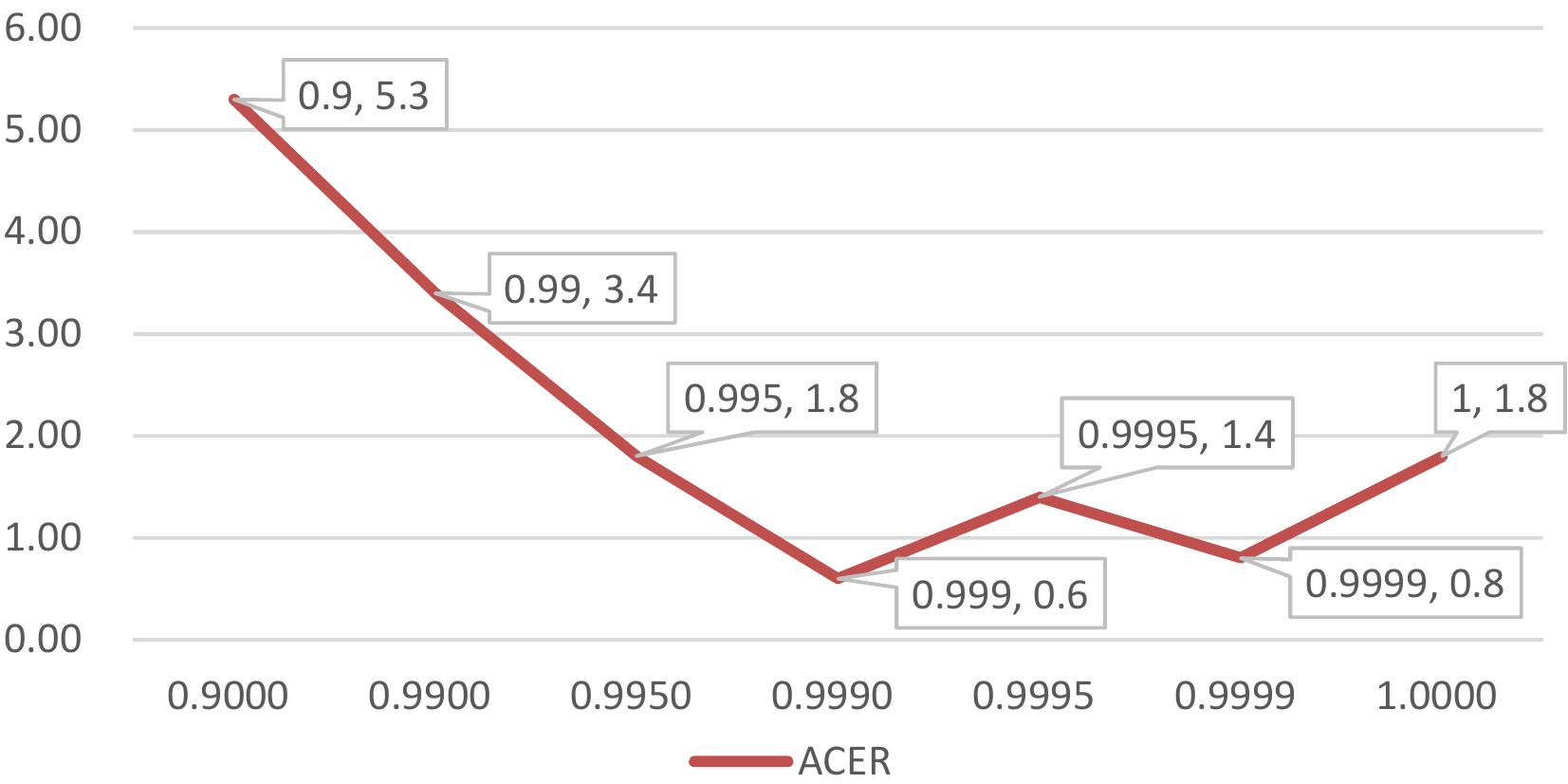}
		\caption{
			The impact of the parameter $\gamma$ towards the meta-teacher and the trained FAS-DR detector.
			The $x$-axis denotes the value of $\gamma$ and the $y$-axis denotes $ACER$ on protocol 1 of OULU-NPU.
		}
		\label{fig:gamma_ablation}
	\end{figure}

	Fig.~\ref{fig:gamma_ablation} shows how $\gamma$ affects the performance of the meta-teacher and further affects the trained detector.
	When $\gamma=0.999$, the detector achieves the best performance with the lowest $ACER$.
	Either smaller or larger $\gamma$ damages the meta-teacher and consequently decreases the FAS-DR detector's performance.
	For instance, compared with $\gamma=0.999$, ${\gamma=0.9}$ harms the meta-teacher and greatly rises FAS-DR(MT)'s $ACER$ to 5.3\%.
	This experiment indicates that updating $MT_v$ either too slow or too fast is unfriendly for the meta-teacher $MT_t$'s training.
	
	\subsubsection{Adapt the surrogate detector to the changing $MT_t$}
	\label{sec:update_detector}
	Within the bi-level optimizing progress of the meta-teacher $MT_t$, we also update the surrogate PA detector to adapt it to the changing $MT_t$.
	Line 12 in Algorithm~\ref{algorithm:Meta-teacher} shows that we use the weight $\theta^*(\omega)$ to update the surrogate PA detector's weight $\theta$.
	In this ablation experiment, we remove Line 12 of Algorithm\ref{algorithm:Meta-teacher} to evaluate whether the update of the surrogate PA detector is necessary to improve the meta-teacher's optimization.
	We denote the newly trained meta-teacher as MT\_w/o\_adapt and denote the FAS-DR detector supervised by MT\_w/o\_adapt as FAS-DR(MT\_w/o\_adapt).
	
	Table \ref{tab:Ablation} reports the performance of FAS-DR(MT\_w/o\_adapt).
	Clearly, FAS-DR(MT) outperforms FAS-DR(MT\_w/o\_adapt), revealing the importance of adapting the surrogate PA detector to the changing $MT_t$.
	One possible underlying reason is that ignoring updating the surrogate detector (without Line 12 of Algorithm~\ref{algorithm:Meta-teacher}) results in a larger representation gap between $MT_t$ and the surrogate detector in the meta-teacher $MT_t$'s training procedure.
	An excessively large representation gap between $MT_t$ and the surrogate detector hinders $MT_t$ from effectively teaching the surrogate detector and results in incorrect evaluation and optimization of $MT_t$.
	
	\begin{figure}[]
		\centering
		\includegraphics[width=0.99\columnwidth]{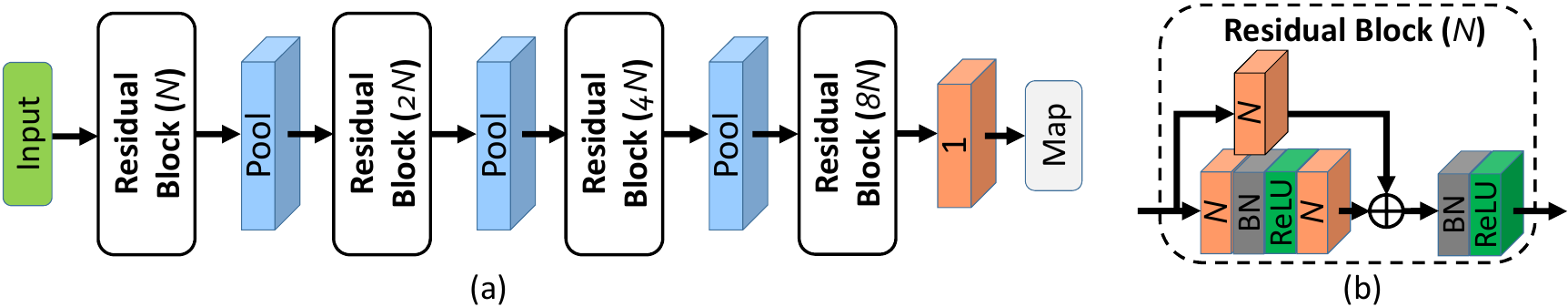}
		\caption{
			(a) The architecture of the ResNet-8 backbone. 
			(b) The inner structure of the residual block.
			Pool: Maxpooling with stride of 2 and pooling size of 2.
			Orange cube: convolution layer.
			BN: batch-normalization.
			Note that the number on orange cube denotes the number of filters of the convolution layer.
			The convolution layer on the shortcut path uses 1$\times$1 stride while all the other convolution layers use 3$\times$3 stride.
			$N$ is set to 16 in this work.
		}
		\label{fig:resnet8}
	\end{figure}

	\subsubsection{Backbone of $MT_t$}
	We set the backbone of $MT_t$ to FAS-DR-Light by default.
	In this experiment, we verify whether the meta-teacher is generalizable to other backbones by replacing its backbone with another backbone called ResNet-8.
	The architecture of ResNet-8 is shown in Fig.~\ref{fig:resnet8}.
	It uses four residual blocks\cite{He2016} to extract features and one convolution layer to regress the map with 32x32 resolution.
	
	We denote the meta-teacher using ResNet-8 backbone as MT\_resnet and denote the FAS-DR detector supervised by MT\_resnet as FAS-DR(MT\_resnet).
	When training MT\_resnet, we keep all the other experimental settings the same as those of the default meta-teacher.
	Table \ref{tab:Ablation} reports the performance of FAS-DR(MT\_resnet).
	Although FAS-DR(MT\_resnet) performs slightly worse than FAS-DR(MT), it is still evident that FAS-DR(MT\_resnet) outperforms FAS-DR(Depth).
	Compared with FAS-DR(Depth), FAS-DR(MT\_resnet) decreases $ACER$ by approximately 47\%, validating the generalizability of the proposed MT-FAS method across different backbones.

	{\subsubsection{Supervising different PA detectors}
	Tables \ref{tab:OULU}, \ref{tab:SiW-M}, and \ref{tab:Domain generation} demonstrate that the meta-teacher provides superior pixel-wise supervision and improves the performances of different PA detectors including FAS-DR\cite{qin2019learning}, CDCN\cite{yu2020searching2}, DTN\cite{DTN}, and  RFMetaFAS\cite{shao2019regularized}.
	Here we further use the meta-teacher to supervise the other two modified FAS-DR backbones who have different numbers of parameters with the default FAS-DR.
	The default FAS-DR backbone used in the work is shown in Fig.~\ref{fig:FAS-DR-BC}.
	Its first (the bottom) convolution layer contains 16 filters.
	In this experiment, except for the last (the topmost) layer, we modify the FAS-DR backbone by doubling or quadrupling the number of filters of all the other convolution layers.
	We denote the new FAS-DR backbone as FAS-DR$_{32}$ or FAS-DR$_{64}$.
	Either FAS-DR$_{32}$ or FAS-DR$_{64}$ has much more parameters than the default FAS-DR backbone.
	We use the meta-teacher to train them and denote the trained detectors FAS-DR$_{32}$(MT) and FAS-DR$_{64}$(MT).
	The corresponding experimental results shown in Table \ref{tab:Ablation} demonstrate that the meta-teacher's teaching performance is insensitive to the number of parameters in PA detectors.
}

	\subsection{Visualization}
	
	\begin{figure}[]
		\centering
		\includegraphics[width=0.99\columnwidth]{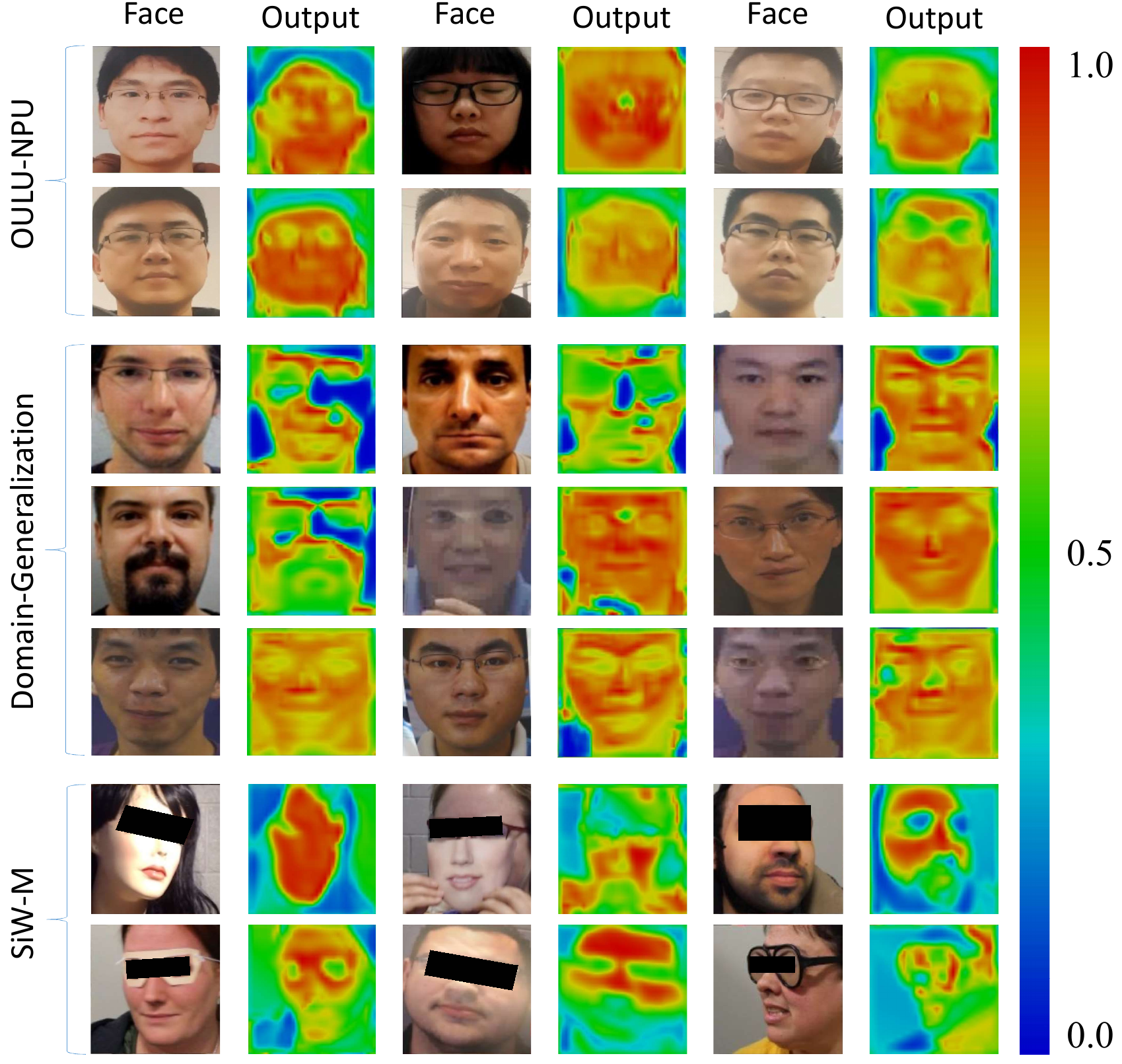}
		\caption{
			The outputs of the meta-teacher for spoof faces from the three benchmarks.
			The colors ranging from blue to red denote the float values from zero to one.
		}
		\label{fig:search_process}
	\end{figure}

	\subsubsection{Prediction of the meta-teacher}
	Fig.~\ref{fig:search_process} visualizes the predictions of the meta-teacher for spoof faces in all three benchmarks.
	The visualization shows that the predictions of the meta-teacher contain facial structure and spoofing cues, especially on SiW-M.
	For example, the prediction for the face in the 6-th row and 3-rd column focuses greatly on the partial mask.
	The prediction for the face in the 7-th row and 5-th column pays more attention to the camouflage glasses, a particular spoof category in SiW-M. 
	Fig.~\ref{fig:search_process} demonstrates that the meta-teacher provides PA detectors with effective pixel-wise supervision that contains rich and intrinsic spoofing cues.
	This finding may explain why the detectors supervised by our meta-teacher outperforms existing vanilla detectors.
	
	\begin{figure}[]
		\centering
		\includegraphics[width=0.99\columnwidth]{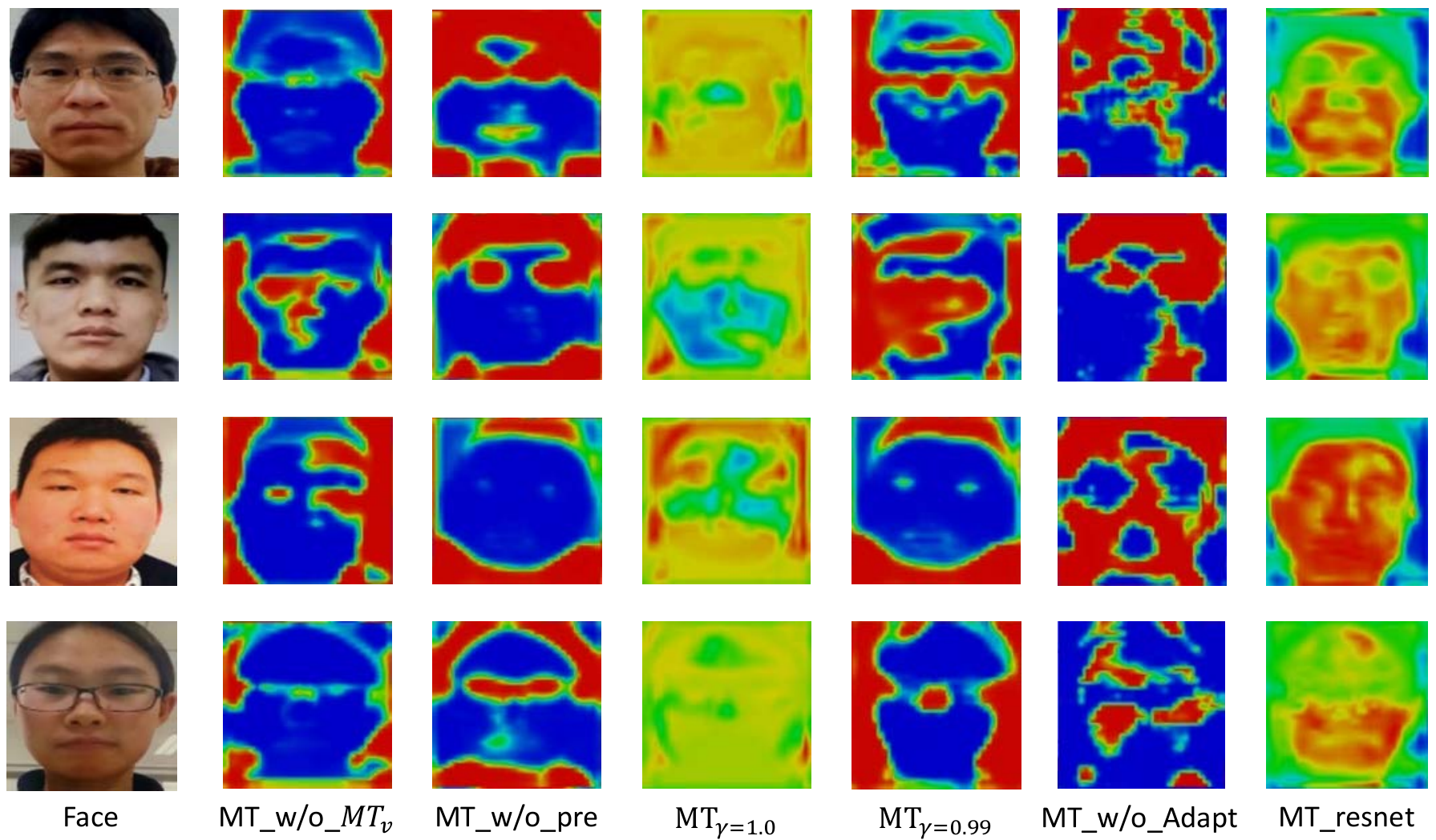}
		\caption{
			The ablation meta-teachers' predictions for spoof faces from OULU-NPU.
			The colors ranging from blue to red denote the float values from zero to one.
		}
		\label{fig:ablation_output}
	\end{figure}

	\subsubsection{Prediction of ablated meta-teachers}
	We also want to know the outputs of the aforementioned ablated meta-teachers.
	Fig.~\ref{fig:ablation_output} visualizes these meta-teachers' outputs for spoof faces from OULU-NPU.
	We note that MT\_w/o\_$MT_v$, MT\_w/o\_pre, MT$_{\gamma=0.99}$, and MT\_w/o\_adapt seem to overfit to some facial regions, which may explain why they could not supervise the detector learning the most precise spoofing cues.
	Setting $\gamma$ to 1.0 will freeze $MT_v$ when training $MT_t$.
	Therefore, MT$_{\gamma=1.0}$ may have difficulty in thoroughly learn the teaching ability.
	Among all the ablated meta-teachers shown in Fig.~\ref{fig:ablation_output}, the output map of MT\_{resnet} contains the richest information.
	This may explain why MT\_{resnet} is more efficient than the other ablated meta-teachers in guiding the PA detector to learn spoofing cues.

\begin{figure}[]
		\centering
		\includegraphics[width=0.99\columnwidth]{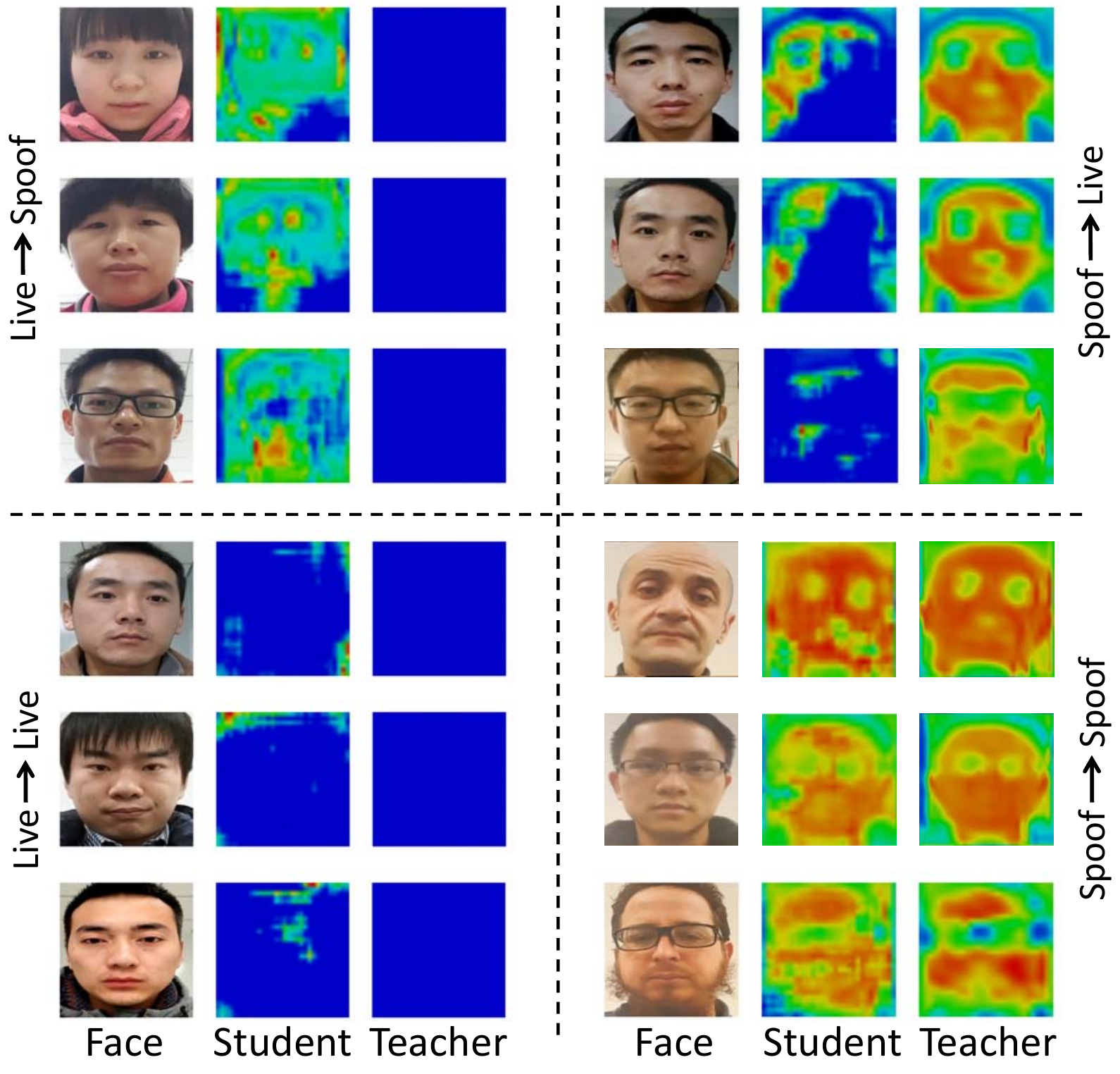}
		\caption{
			Some faces that are correctly and wrongly classified by FAS-DR(MT).
			The `Student' column denotes the detector FAS-DR(MT)'s output for the corresponding face.
			The `Teacher' column denotes the meta-teacher's output.
			The colors ranging from blue to red denote the float numbers from zero to one.
			Blue denotes 0 and Red denotes 1.
		}
		\label{fig:samples}
	\end{figure}

	{
	\subsubsection{Correctly and wrongly classified faces}
	Fig.~\ref{fig:samples} visualizes some testing faces that are correctly and wrongly classified by FAS-DR(MT).
	All visualized faces are sampled from the testing set of protocol 2 in OULU-NPU.
	Live $\rightarrow$ Spoof denotes the live faces that are wrongly classified as spoof faces.
	Spoof $\rightarrow$ Live denotes the spoof faces that are wrongly classified as live faces.
	Live $\rightarrow$ Live and Spoof $\rightarrow$ Spoof denote the correctly classified live and spoof faces, respectively.
	Fig.~\ref{fig:samples} also visualizes the corresponding pixel-wise maps predicted by FAS-DR(MT) for understanding why the faces are correctly or wrongly classified.
	In our work, FAS-DR(MT) classifies each testing face by comparing the average value of the predicted map with the threshold.
	Given the predicted map $\hat{y}_{map}$, the testing face will be classified as a spoof face if $mean(\hat{y}_{map}) \textgreater$threshold, otherwise, it will be classified as a live face.
	For instance, the three live faces in the left upper corner are misclassified as spoof faces because the averages of the predicted maps are larger than the threshold.
	Besides, for illustrating how the meta-teacher teaching the PA detector, we also show the pixel-wise maps outputted by the meta-teacher in Fig.~\ref{fig:samples}.
}

	\section{Conclusion and Future Work}
	\label{sec:conclusion}
	Existing deep learning-based FAS methods use handcrafted labels to supervise the PA detector's learning.
	Although handcrafted labels perform satisfactorily in supervising existing PA detectors, these labels rely heavily on human's prior-knowledge about FAS and may lose effectiveness against newly developed spoof types.
	In this paper, we aim to explore better supervision towards PA detectors from a novel perspective.
	To this end, we propose a novel meta-teacher face anti-spoofing (MT-FAS) method, in which a meta-teacher is trained to learn how to provide better-suited supervision to the PA detector.
	Once the meta-teacher is trained, we use it to supervise existing PA detectors' training and improve these detectors' performance.
	By extensive experiments, we demonstrate that the meta-teacher outperforms not only the most widely employed manually-designed labels but also existing teacher-student methods in training PA detectors.
	Moreover, with the advantage of the meta-teacher, we improve upon the state-of-the-art performances on several popular FAS benchmarks.
	
	{Despite the demonstrated advantages of the trained meta-teacher, we should note that the training of the meta-teacher needs complex second-order gradient, as Eq.~\ref{eq:label_optimize6} shows.
	In the future, we will dive deeper into MT-FAS and try to reduce the requirement of calculating second-order gradient.
	Moreover, extending MT-FAS to other computer vision tasks (\emph{e.g.}, classification, noisy-label problems) is also an interesting direction worthy to be explored.}

	\section*{Acknowledgments}
	This work was supported by the National Key Research and Development Program of China (No. 2020AAA0140002). This work was also supported in part by the National Natural Science Foundation of China (No. 61876178, 61976229).
	\ifCLASSOPTIONcaptionsoff
	\newpage
	\fi

	
	
	%
	\bibliographystyle{IEEEtran}
	\bibliography{IEEEabrv,main}

	
	
\end{document}